\documentclass{article}

\PassOptionsToPackage{numbers, sort&compress}{natbib}

\usepackage[final]{neurips_2022}

\usepackage[utf8]{inputenc} %
\usepackage[T1]{fontenc}    %
\usepackage{hyperref}       %
\usepackage{url}            %
\usepackage{booktabs}       %
\usepackage{amsfonts}       %
\usepackage{nicefrac}       %
\usepackage{microtype}      %
\usepackage{amsmath}
\usepackage[dvipsnames]{xcolor}
\usepackage{graphicx}
\usepackage{amssymb}
\usepackage{cancel}
\usepackage{algorithm}
\usepackage[noend]{algpseudocode}
\usepackage[frozencache, cachedir=.]{minted}
\usepackage{subcaption}
\usepackage{bbm}
\usepackage{amsthm}
\usepackage{wrapfig}
\usepackage{tcolorbox}

\usepackage{makecell}
\usepackage{amssymb}%
\usepackage{pifont}%
\usepackage{multicol}

\usepackage{amsmath,amsfonts,bm}

\newcommand{\figtop}{{\em (Top)}}
\newcommand{\figbottom}{{\em (Bottom)}}

\def\eqref#1{equation~\ref{#1}}

\def\1{\bm{1}}

\DeclareMathAlphabet{\mathsfit}{\encodingdefault}{\sfdefault}{m}{sl}
\SetMathAlphabet{\mathsfit}{bold}{\encodingdefault}{\sfdefault}{bx}{n}

\def\gL{{\mathcal{L}}}

\def\gS{{\mathcal{S}}}

\newcommand{\E}{\mathbb{E}}

\newtheorem{theorem}{Theorem}[section]

\newtheorem{lemma}[theorem]{Lemma}

\newtheorem{proposition}{Proposition}

\title{Contrastive Learning as Goal-Conditioned Reinforcement Learning}

\author{%
Benjamin Eysenbach$^{\alpha,\beta}$ \quad Tianjun Zhang$^\gamma$ \quad Sergey Levine$^{\beta,\gamma}$ \quad  Ruslan Salakhutdinov$^{\alpha}$ \\
$^\alpha$CMU \qquad $^\beta$Google Research \qquad $^\gamma$UC Berkeley
}

\begin{document}
\usemintedstyle{tango}

\maketitle

\begin{abstract}

In reinforcement learning (RL), it is easier to solve a task if given a good representation. While \emph{deep} RL should automatically acquire such good representations, prior work often finds that learning representations in an end-to-end fashion is unstable and instead equip RL algorithms with additional representation learning parts (e.g., auxiliary losses, data augmentation). How can we design RL algorithms that directly acquire good representations? In this paper, instead of adding representation learning parts to an existing RL algorithm, we show (contrastive) representation learning methods can be cast as RL algorithms in their own right. To do this, we build upon prior work and apply contrastive representation learning to action-labeled trajectories, in such a way that the (inner product of) learned representations exactly corresponds to a goal-conditioned value function. We use this idea to reinterpret a prior RL method as performing contrastive learning, and then use the idea to propose a much simpler method that achieves similar performance. Across a range of goal-conditioned RL tasks, we demonstrate that contrastive RL methods achieve higher success rates than prior non-contrastive methods, including in the offline RL setting. We also show that contrastive RL outperforms prior methods on image-based tasks, without using data augmentation or auxiliary objectives. \footnote{Project website with videos and code: \url{https://ben-eysenbach.github.io/contrastive_rl}}

\end{abstract}

\section{Introduction}

Representation learning is an integral part of reinforcement learning (RL\footnote{RL = reinforcement learning, not representation learning.}) algorithms. While such representations might emerge from end-to-end training~\citep{wang2022investigating, liang2015state, such2018atari, annasamy2019towards}, prior work has found it necessary to equip RL algorithms with perception-specific loss functions~\citep{srinivas2020curl, nachum2018near, guo2018neural, lange2010deep, finn2016deep, nair2018visual, Rakelly2021WhichMR, zhang2020learning} or data augmentations~\citep{srinivas2020curl, stooke2021decoupling, laskin2020reinforcement, kostrikov2020image}, effectively decoupling the representation learning problem from the reinforcement learning problem. 
Given what prior work has shown about RL in the presence of function approximation and state aliasing~\citep{yu2020gradient, yang2021overcoming, achiam2019towards}, it is not surprising that end-to-end learning of representations is fragile~\citep{laskin2020reinforcement, kostrikov2020image}: an algorithm needs good representations to drive the learning of the RL algorithm, but the RL algorithm needs to drive the learning of good representations. So, \emph{can we design RL algorithms that do learn good representations without the need for auxiliary perception losses?}

\begin{figure}
    \centering
    \vspace{-2em}
    \includegraphics[width=0.85\linewidth]{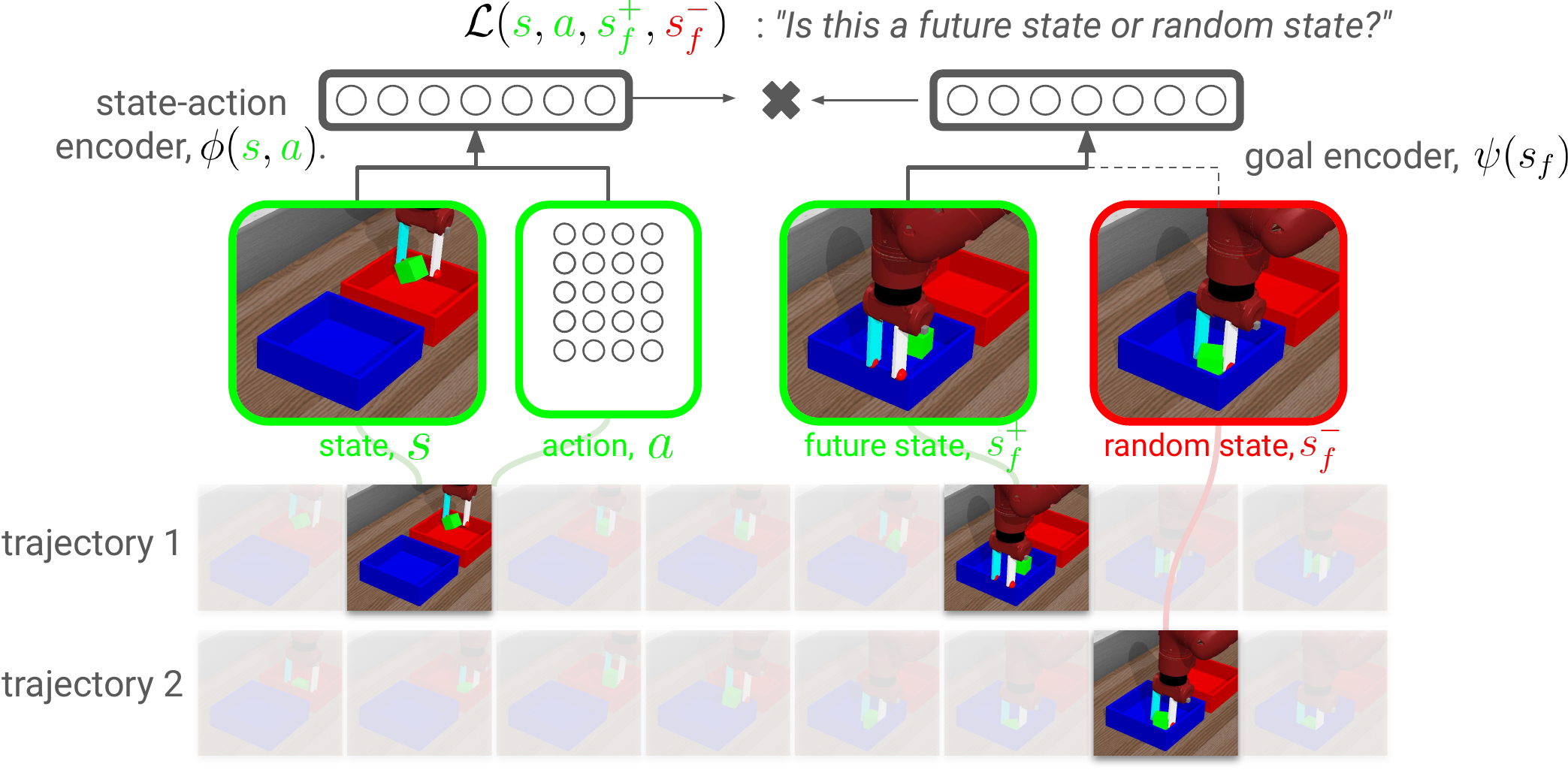}
    \caption{
    \footnotesize
    \textbf{Reinforcement learning via contrastive learning.}
    Our method uses contrastive learning to acquire representations of state-action pairs ($\phi(s, a)$) and future states ($\psi(s_f)$), so that the representations of future states are closer than the representations of random states. We prove that learned representation corresponds to a value function for a certain reward function. To select actions for reaching goal $s_g$, the policy chooses the action where $\phi(s, a)$ is closest to $\psi(s_g)$.}
    \vspace{-1em}
    \label{fig:method}
\end{figure}

Rather than using a reinforcement learning algorithm also to solve a representation learning problem, we will use a representation learning algorithm to also solve certain types of reinforcement learning problems, namely goal-conditioned RL. Goal-conditioned RL is widely studied~\citep{chane2021goal, ding2019goal, sun2019policy, andrychowicz2017hindsight, kaelbling1993learning, Lin2019ReinforcementLW}, and intriguing from a representation learning perspective because it can be done in an entirely self-supervised manner, without manually-specified reward functions.
We will focus on contrastive (representation) learning methods, using observations from the same trajectory (as done in prior work~\citep{oord2018representation, sermanet2018time}) while also including actions as an additional input (See Fig.~\ref{fig:method}). Intuitively, contrastive learning then resembles a goal-conditioned value function: nearby states have similar representations and unreachable states have different representations. We make this connection precise, showing that sampling positive pairs using the discounted state occupancy measure results in learning representations whose inner product exactly corresponds to a value function.

In this paper, we show how contrastive representation learning can be used to perform goal-conditioned RL. We formally relate the learned representations to reward maximization, showing that the inner product between representations corresponds to a value function. This framework of contrastive RL generalizes prior methods, such as C-learning~\citep{Eysenbach2021CLearningLT}, and suggests new goal-conditioned RL algorithms. One new method achieves performance similar to prior methods but is simpler; another method consistently outperforms the prior methods. On goal-conditioned RL tasks with image observations, contrastive RL methods outperform prior methods that employ data augmentation and auxiliary objectives, and do so without data augmentation or auxiliary objectives.
In the offline setting, contrastive RL can outperform prior methods on benchmark goal-reaching tasks, sometimes by a wide margin.

\section{Related Work}
\label{sec:prior-work}

This paper will draw a connection between RL and contrastive representation learning, building upon a long line of contrastive learning methods in NLP and computer vision, and deep metric learning~\citep{ma2018noise, hjelm2018learning, mnih2012fast,levy2014neural,  mikolov2013distributed,nowozin2016f,Sohn2016ImprovedDM, oord2018representation, Wu2018UnsupervisedFL, hjelm2018learning, he2020momentum, Chen2020ASF, tian2020contrastive, Weinberger2005DistanceML, Schroff2015FaceNetAU, hoffer2015deep, sermanet2018time}. Contrastive learning methods learn representations such that similar (``positive'') examples have similar representations and dissimilar (``negative'') examples have dissimilar representations.\footnote{Our focus will not be on recent methods that learn representations without negative samples~\citep{Grill2020BootstrapYO, Chen2021ExploringSS}.}
While most methods generate the ``positive'' examples via data augmentation, some methods generate similar examples using different camera viewpoints of the same scene~\citep{sermanet2018time, tian2020contrastive}, or by sampling examples that occur close in time within time series data~\citep{oord2018representation, sermanet2018time, anand2019unsupervised, stooke2021decoupling}. Our analysis will focus on this latter strategy, as the dependence on time will allow us to draw a precise relationship with the time dependence in RL.

\emph{Deep} RL algorithms promise to automatically learn good representations, in an end-to-end fashion. However, prior work has found it challenging to uphold this promise~\citep{wang2022investigating, liang2015state, such2018atari, annasamy2019towards}, prompting many prior methods to employ separate objectives for representation learning and RL~\citep{srinivas2020curl, nachum2018near, guo2018neural, lange2010deep, finn2016deep, nair2018visual, Rakelly2021WhichMR, zhang2020learning, stooke2021decoupling,zhang2022making, qiu2022contrastive}.  
Many prior methods choose a representation learning objectives that reconstruct the input state~\citep{lange2010deep, finn2016deep, ha2018world, Nasiriany2019PlanningWG, nair2018visual, zhang2019solar, hafner2019learning, hafner2019dream} while others use contrastive representation learning methods~\citep{stooke2021decoupling, srinivas2020curl, nachum2018near, oord2018representation, shu2020predictive}.
Unlike these prior methods, we will not use a separate representation learning objective, but instead use the same objective for both representation learning and reinforcement learning.
Some prior RL methods have also used contrastive learning to acquire reward functions~\citep{fischinger2013learning, christiano2017deep, xie2018few, Fu2018VariationalIC, brown2019extrapolating, zolna2019task, xu2019positive, konyushkova2020semi, Kalashnikov2021MTOptCM, nair2022learning}, often in imitation learning settings~\citep{ho2016generative, fu2017learning}. In contrast, we will use contrastive learning to directly acquire a value function, which (unlike a reward function) can be used directly to take actions, without any additional RL.

This paper will focus on goal-conditioned RL problems, a problem prior work has approached using temporal difference learning~\citep{kaelbling1993learning, schaul2015universal, andrychowicz2017hindsight,Lin2019ReinforcementLW, Eysenbach2021CLearningLT, rudner2021outcome}, conditional imitation learning~\citep{savinov2018semi, ding2019goal, sun2019policy, ghosh2020learning, lynch2020learning}, model-based methods~\citep{dosovitskiy2016learning, schmeckpeper2020learning}, hierarchical RL~\citep{nachum2018data}, and planning-based methods~\citep{savinov2018semi, Srinivas2018UniversalPN, Nasiriany2019PlanningWG, eysenbach2019search}.
The problems of automatically sampling goals and exploration~\citep{du2021curious, pong2019skew, florensa2018automatic, zhao2019maximum, mendonca2021discovering} are orthogonal to this work. Like prior work, we will parametrize the  value function as an inner product between learned representations~\citep{hong2022bilinear, schaul2015universal, florensa2019self}. Unlike these prior methods, we will learn a value function directly via contrastive learning, without using reward functions or TD learning.

Our analysis will be most similar to prior methods~\citep{chane2021goal, Eysenbach2021CLearningLT, blier2021learning, rudner2021outcome} that view goal-conditioned RL as a data-driven problem, rather than as a reward-maximization problem.
Many of these methods employ hindsight relabeling~\citep{kaelbling1993learning, andrychowicz2017hindsight, Eysenbach2020RewritingHW, li2020generalized}, wherein experience is \emph{relabeled} with an outcome that occurred in the future. Whereas hindsight relabeling is typically viewed as a trick to add on top of an RL algorithm, this paper can roughly be interpreted as showing that the hindsight relabeling is a standalone RL algorithm. Many goal-conditioned methods learn a value function that captures the similarity between two states~\citep{kaelbling1993learning, nair2018visual, Venkattaramanujam2019SelfsupervisedLO, Eysenbach2021CLearningLT}. Such distance functions are structurally similar to the critic function learned for contrastive learning, a connection we make precisely in Sec.~\ref{sec:method}. In fact, our analysis shows that C-learning~\citep{Eysenbach2021CLearningLT} is already performing contrastive learning, and our experiments show that alternative contrastive RL methods can be much simpler and achieve higher performance.

Prior work has studied how representations related to reward functions using the framework of universal value functions~\citep{schaul2015universal, borsa2018universal} and successor features~\citep{liu2021aps, barreto2017successor, hansen2019fast}.
While these methods typically require additional supervision to drive representation learning (manually-specified reward functions or features), our method is more similar to prior work that estimates the discounted state occupancy measure as an inner product between learned representations~\citep{blier2021learning,wu2018laplacian}.
While these methods use temporal difference learning, ours is akin to Monte Carlo learning. While Monte Carlo learning is often (but not always~\citep{dosovitskiy2016learning}) perceived as less sampling efficient, our experiments find that our approach can be as sample efficient as TD methods.
Other prior work has focused on learning representations that can be used for planning~\citep{watter2015embed, ichter2020broadly, liu2020hallucinative, rybkin2021model, savinov2018semi}. Our method will learn representations using an objective similar to prior work~\citep{sermanet2018time, savinov2018semi}, but makes the key observation that the representation already encodes a value function: no additional planning or RL is necessary to choose actions.

Please see Appendix~\ref{appendix:related} for a discussion of how our work relates to unsupervised skill learning.

\section{Preliminaries}
\label{sec:prelims}

\textbf{Goal-conditioned reinforcement learning.}
The goal-conditioned RL problem is defined by states $s_t \in \gS$, actions $a_t$, an initial state distribution $p_0(s)$, the dynamics $p(s_{t+1} \mid s_t, a_t)$, a distribution over goals $p_g(s_g)$, and a reward function $r_g(s, a)$ for each goal.
This problem is equivalent to a multi-task RL~\citep{teh2017distral, wilson2007multi, andreas2017modular, guo2020bootstrap, yu2020meta}, where tasks correspond to reaching goals states. Following prior work~\citep{Eysenbach2021CLearningLT, chane2021goal, blier2021learning, rudner2021outcome}, we define the reward as the probability (density) of reaching the goal at the next time step:\footnote{At the initial state, this reward also includes the probability that the agent started at the goal: $r_g(s_0, a_0) = (1 - \gamma)(p(s_1 = s_g \mid s_0, a_0) + p_0(s_0 = s_g))$}
{\footnotesize \begin{align}
    r_g(s_t, a_t) \triangleq (1 - \gamma) p(s_{t+1} = s_g \mid s_t, a_t). \label{eq:reward}
\end{align}}\!\!
This reward function is appealing because it avoids the need for a human user to specify a distance metric (unlike, e.g.,~\citep{andrychowicz2017hindsight}). Even though our method will not estimate the reward function, we will still use the reward function for analysis.
For a goal-conditioned policy $\pi(a \mid s, s_g)$, we
use $\pi(\tau \mid s_g)$ to denote the probability of sampling an infinite-length trajectory $\tau = (s_0, a_0, s_1, a_1, \cdots)$. We defined the expected reward objective and Q-function as
{\footnotesize \begin{align}
    \max_{\pi} 
    \E_{p_g(s_g), \pi(\tau \mid s_g)}\left[ \sum_{t=0}^\infty \gamma^t r_g(s_t, a_t) \right], \quad
    Q_{s_g}^\pi(s, a) \triangleq \E_{\pi(\tau \mid s_g)}\left[\sum_{t' = t}^\infty \gamma^{t' - t} r_g(s_{t'}, a_{t'}) \mid \substack{s_t = s,\\a_t = a} \right]. \label{eq:q}
\end{align}}\!\!
Intuitively, this objective corresponds to sampling a goal $s_g$ and then optimizing the policy to go to that goal and stay there.
Finally, we define the discounted state occupancy measure as~\citep{ho2016generative, zhang2020gradientdice}
{\footnotesize \begin{equation}
    p^{\pi(\cdot \mid \cdot, s_g)} (s_{t+} = s) \triangleq (1 - \gamma) \sum_{t=0}^\infty \gamma^t p_t^{\pi(\cdot \mid \cdot, s_g)}(s_t = s), \label{eq:future-states}
\end{equation}}\!\!
where $p_t^\pi(s)$ is the probability density over states that policy $\pi$ visits after $t$ steps. Sampling from the discounted state occupancy measure is easy: the first sample a time offset from a geometric distribution ($t \sim \textsc{Geom}(1 - \gamma)$), and then look at what state the policy visits after exactly $t$ steps. We will use $s_{t+}$ to denote states sampled from the discounted state occupancy measure. Because our method will combine experience collected from multiple policies, we also define the average stationary distribution as $p^{\pi(\cdot \mid \cdot)}(s_{t+} = s \mid s, a) \triangleq \int p^{\pi(\cdot \mid \cdot, s_g)}(s_{t+} = s \mid s, a) p^\pi(s_g \mid s, a)  ds_g$, where $p^\pi(s_g \mid s, a)$ is the probability of the \emph{commanded} goal given the current state-action pair. This stationary distribution is equivalent to that of the policy $\pi(a \mid s) \triangleq \int \pi(a \mid s, s_g) p^\pi(s_g \mid s) ds_g$~\citep{ziebart2010modeling}.

\textbf{Contrastive representation learning.}
Contrastive representation learning methods~\citep{Weinberger2005DistanceML, gutmann2012noise, mnih2012fast, mikolov2013distributed, levy2014neural, jozefowicz2016exploring, hjelm2018learning, ma2018noise, tschannen2019mutual, he2020momentum, tian2020contrastive, Chen2020ASF} take as input pairs of positive and negative examples, and learn representations so that positive pairs have similar representations and negative pairs have dissimilar representations. We use $(u, v)$ to denote an input pair (e.g., $u$ is an image, and $v$ is an augmented version of that image). Positive examples are sampled from a joint distribution $p(u, v)$, while negative examples are sampled from the product of marginal distributions, $p(u)p(v)$.
We will use an objective based on binary classification~\citep{mnih2012fast,levy2014neural,  mikolov2013distributed,nowozin2016f}. Let $f(u, v) = \phi(u)^T\psi(v)$ be the similarity between the representations of $u$ and $v$. We will call $f$ the \emph{critic function}\footnote{In contrastive learning, the critic function indicates the similarity between a pair of inputs~\citep{poole2019variational}; in RL, the critic function indicates the future expected returns~\citep{konda1999actor}. Our method combines contrastive learning and RL in a way that these meanings become one and the same.}
and note that its range is $(-\infty, \infty)$.
We will use NCE-binary~\citep{ma2018noise} objective (also known as InfoMAX~\citep{hjelm2018learning}):
{\footnotesize \begin{equation}
    \max_{f(u, v)} \E_{\substack{(u, {\color{OliveGreen}v^+}) \sim p(u, v)\\{\color{red}v^-} \sim p(u)}}\bigg[ \log \sigma(\underbrace{f(u, {\color{OliveGreen}v^+})}_{\phi(u)^T\psi({\color{OliveGreen}v^+})}) + \log(1 - \sigma(\underbrace{f(u, {\color{red}v^-})}_{\phi(u)^T\psi({\color{red}v^-})})) \bigg]. \label{eq:nce-orig}
\end{equation}}\!\!

\section{Contrastive Learning as an RL Algorithm}
\label{sec:method}

This section shows how to use contrastive representation to \emph{directly} perform goal-conditioned RL. The key idea (Lemma~\ref{lemma:critic-is-q}) is that contrastive learning estimates the Q-function for a certain policy and reward function.
To prove this result, we relate the Q-function to the state occupancy measure (Sec.~\ref{sec:q-probs}) and then relate the optimal critic function to the state occupancy measure (Sec.~\ref{sec:opt-critic}).

This result allows us to propose a new algorithm for goal-conditioned RL based on contrastive learning. Unlike prior work, this algorithm is not adding contrastive learning on top of an existing RL algorithm. This framework generalizes C-learning~\citep{Eysenbach2021CLearningLT}, offering a cogent explanation for its good performance while also suggesting new methods that are simpler and can achieve higher performance.

\subsection{Relating the Q-function to probabilities}
\label{sec:q-probs}
This section sets the stage for the main results of this section by providing a probabilistic perspective goal-conditioned RL. The expected reward objective and associated Q-function in (Eq.~\ref{eq:q}) can equivalently be expressed as the probability (density) of reaching a goal in the future:
\begin{proposition}[rewards $\rightarrow$ probabilities] \label{prop:q}
The Q-function for the goal-conditioned reward function $r_g$ (Eq.~\ref{eq:reward}) is equivalent to the probability of state $s_g$ under the discounted state occupancy measure:
\begin{equation}
    Q_{s_g}^\pi(s, a) = p^{\pi(\cdot \mid \cdot, s_g)}(s_{t+} = s_g \mid s, a).
\end{equation}
\end{proposition}
The proof is in Appendix~\ref{appendix:proofs}.
Translating rewards into probabilities not only makes it easier to analyze the goal-conditioned problem, but also means that any method for estimating probabilities (e.g., contrastive learning) can be turned into a method for estimating this Q-function. %

\subsection{Contrastive Learning Estimates a Q-Function}
\label{sec:opt-critic}

We will use contrastive learning to learn a value function by carefully choosing the inputs $u$ and $v$. The first input, $u$, will correspond to a state-action pair, $u = (s_t, a_t) \sim p(s, a)$. In practice, these pairs are sampled from the replay buffer.
Including the actions in the input is important because it will allow us to determine which actions to take to reach a desired future state. The second variable, $v$, is a \emph{future} state, $v = s_f$. For the ``positive'' training pairs, the future state is sampled from the discounted state occupancy measure, $s_f \sim p^{\pi(\cdot \mid \cdot)}(s_{t+} \mid s_t, a_t)$. For the ``negative'' training pairs, we sample a future state from a random state-action pair: $s_f \sim p(s_{t+}) \triangleq \int p^{\pi(\cdot \mid \cdot)}(s_{t+} \mid s, a)p(s, a) ds da$.
With these inputs, the contrastive learning objective (Eq.~\ref{eq:nce-orig}) can be written as
{\footnotesize \begin{align}
   & \max_f \E_{\substack{(s, a) \sim p(s, a),{\color{red}s_f^-} \sim p(s_f) \\
   {\color{green}s_f^+} \sim p^{\pi(\cdot \mid \cdot)}(s_{t+} \mid s_t, a_t)}}\left[\gL(s, a, {\color{green}s_f^+}, {\color{red}s_f^-}) \right], \nonumber  \\
    & \qquad \text{where} \quad  \gL(s, a, {\color{green}s_f^+}, {\color{red}s_f^-}) \triangleq \log \sigma(\underbrace{f(s, a, {\color{green}s_f^+})}_{\phi(s, a)^T\psi({\color{green}s_f^+})}) + \log (1 - \sigma(\underbrace{f(s, a, {\color{red}s_f^-})}_{\phi(s, a)^T\psi({\color{red}s_f^-})}) ).
    \label{eq:nce-rl}
\end{align}}\!\!
Intuitively, the critic function $f(u = (s_t, a_t), v = s_f)$ now tells us the correlation between the current state-action pair and future outcomes, analogous to a Q-function. We therefore can use the critic function in the same way as actor-critic RL algorithms~\citep{konda1999actor}, figuring out which actions lead to the desired outcome. 
Because the Bayes-optimal critic function is a function of the state occupancy measure~\citep{ma2018noise}, $f^*(s, a, s_g) = \log \left( \frac{p^{\pi(\cdot \mid \cdot)}(s_{t+} = s_g \mid s, a)}{p(s_g)} \right)$,
it can be used to express the Q-function:
\begin{lemma} \label{lemma:critic-is-q}
The critic function that optimizes Eq.~\ref{eq:nce-rl} is a Q-function for the goal-conditioned reward function (Eq.~\ref{eq:reward}), up to a multiplicative constant $\frac{1}{p(s_f)}$: $\exp(f^*(s, a, s_f)) = \frac{1}{p(s_f)} \cdot Q_{s_f}^{\pi(\cdot \mid \cdot)}(s, a)$. 
\end{lemma}
The critic function can be viewed as an unnormalized density model, where $p(s_g)$ is the partition function. Much of the appeal of contrastive learning is it avoids estimating the partition function~\citep{gutmann2012noise}, which can be challenging; in the RL setting, it will turn out that this constant can be ignored when selecting actions. Our experiments show that learning a normalized density model works well when $s_g$ is low-dimensional, but struggles to solve higher-dimensional tasks.

This lemma relates the critic function to $Q_{s_f}^{{\color{blue}\pi(\cdot \mid \cdot)}}(s, a)$, not $Q_{s_f}^{{\color{blue}\pi(\cdot \mid \cdot, s_f)}}(s, a)$. The underlying reason is that the critic function combines together experience collected when commanding different goals. Prior goal-conditioned behavioral cloning methods~\citep{ding2019goal, ghosh2020learning, sun2019policy, lynch2020learning} perform similar sharing, but do not analyze the relationship between the learned policies and Q functions. Sec.~\ref{sec:analysis} shows that this critic function can be used as the basis for a convergent RL algorithm under some assumptions.

\subsection{Learning the Goal-Conditioned Policy}
The learned critic function not only tells us the likelihood of future states, but also tells us how different actions change the likelihood of a state occurring in the future. Thus, to learn a policy for reaching a goal state, we choose the actions that make that state most likely to occur in the future:
{\footnotesize \begin{equation}
    \max_{\pi(a \mid s, s_g)} \E_{\pi(a \mid s, s_g)p(s)p(s_g)}\left[f(s, a, s_f = s_g) \right] \approx  \E_{\pi(a \mid s, s_g)p(s)p(s_g)}\left[\log Q_{s_g}^{\pi(\cdot \mid \cdot)}(s, a) - \log p(s_g) \right]. \label{eq:actor-loss}
\end{equation}}\!\!
The approximation above reflects errors in learning the optimal critic, and will allow us to prove that this policy loss corresponds to policy improvement in Sec.~\ref{sec:analysis}, under some assumptions.

In practice, we parametrize the goal-conditioned policy as a neural network that takes as input the state and goal and outputs a distribution over actions. The actor loss (Eq.~\ref{eq:actor-loss}) is computed by sampling states and \emph{random} goals from the replay buffer, sampling actions from the policy, and then taking gradients on the policy using a reparametrization gradient.
On tasks with image observations, we add an action entropy term to the policy objective.

\subsection{A Complete Goal-Conditioned RL Algorithm}
\label{sec:complete}

The complete algorithm alternates between fitting the critic function using contrastive learning, updating the policy using Eq.~\ref{eq:actor-loss}, and collecting more data. Alg.~\ref{alg:contrastive} provides a JAX~\citep{jax2018github} implementation of the actor and critic losses.
Note that the critic is parameterized as an inner product between a representation of the state-action pair, and a representation of the goal state: $f(s, a, s_g) = \phi(s, a)^T\psi(s_g)$. This parameterization allows for efficient computation, as we can compute the goal representations just once, and use them both in the positive pairs and the negative pairs. While this is common practice in representation learning, it is not exploited by most goal-conditioned RL algorithms. We refer to this method as \texttt{contrastive RL (NCE)}. In Appendix~\ref{appendix:cpc}, we derive a variant of this method (\texttt{contrastive RL (CPC)}) that uses the infoNCE bound on mutual information.

Contrastive RL (NCE) is an on-policy algorithm because it only estimates the Q-function for the policy that collected the data. However, in practice, we take as many gradient steps on each transition as standard off-policy RL algorithms~\citep{haarnoja2018soft, fujimoto2018addressing}.
Please see Appendix~\ref{appendix:hparams} for full implementation details. We will also release an efficient implementation based on ACME~\citep{hoffman2020acme} and JAX~\citep{jax2018github}. On a single TPUv2, training proceeds at $1100 \frac{\text{batches}}{\text{sec}}$ for state-based tasks and $105 \frac{\text{batches}}{\text{sec}}$ for image-based tasks; for comparison, our implementation of DrQ on the same hardware setup runs at $28  \frac{\text{batches}}{\text{sec}}$ ($3.9\times$ slower).\footnote{The more recent DrQ-v2~\citep{yarats2021mastering} uses on 1 NVIDIA V100 GPU to achieve a training speed of $96/4 = 24 \frac{\text{batches}}{\text{sec}}$. The factor of 4 comes from an action repeat of 2 and an update interval of 2.}
Architectures and hyperparameters are described in Appendix~\ref{appendix:hparams}.\footnote{Code and more results are available: \url{https://ben-eysenbach.github.io/contrastive_rl}}

\begin{figure}[t]
\vspace{-2em}
\begin{algorithm}[H]
\footnotesize
\caption{\textbf{Contrastive RL (NCE)}: the actor and critic losses for our method.}  \label{alg:contrastive}
\begin{minted}{python}
from jax.numpy import einsum, eye
from optax import sigmoid_binary_cross_entropy
def critic_loss(states, actions, future_states):
  sa_repr = sa_encoder(states, actions)  # (batch_dim, repr_dim)
  g_repr = g_encoder(future_states)      # (batch_dim, repr_dim)
  logits = einsum('ik,jk->ij', sa_repr, g_repr)  # <sa_repr[i], g_repr[j]> for all i,j
  return sigmoid_binary_cross_entropy(logits=logits, labels=eye(batch_size))

def actor_loss(states, goals):
  actions = policy.sample(states, goal=goals)  # (batch_size, action_dim)
  sa_repr = sa_encoder(states, actions)        # (batch_dim, repr_dim)
  g_repr = g_encoder(goals)                    # (batch_dim, repr_dim)
  logits = einsum('ik,ik->i', sa_repr, g_repr)  # <sa_repr[i], g_repr[i]>
  return -1.0 * logits
\end{minted}
\end{algorithm}
\vspace{-2em}
\end{figure}

\subsection{Convergence Guarantees}
\label{sec:analysis}

In general, providing convergence guarantees for methods that perform relabeling is challenging. Most prior work offers no guarantees~\citep{dosovitskiy2016learning, ding2019goal, andrychowicz2017hindsight} or guarantees under only restrictive assumptions~\citep{ghosh2020learning, sun2019policy}.

To prove that contrastive RL converges, we will introduce an additional filtering step into the method, throwing away some training examples.
Precisely, we exclude training examples $(s, a, s_f)$ if the probability of the corresponding trajectory $\tau_{i:j} = (s_i, a_i, s_{i+1}, a_{i+1}, \cdots, s_j, a_j)$ sampled from $\pi(\tau \mid s_g)$ under the commanded goal $s_g$ is very different from the trajectory's probability under the actually-reached goal $s_j$:
{\footnotesize \begin{equation*}
\textsc{ExcludeTraj}(\tau_{i:j}) = \delta\left(\left| \frac{\pi(\tau_{i:j} \mid s_g)}{\pi(\tau_{i:j} \mid s_j)} - 1 \right| > \epsilon\right).
\end{equation*}}\!\!
While this modification is necessary to prove convergence, ablation experiments in Appendix Fig.~\ref{fig:filtering} show that the filtering step can actually hurt performance in practice, so we do not include this filtering step in the experiments in the main text.  We can now prove that contrastive RL performs approximate policy improvement.
\begin{lemma}[Approximate policy improvement] \label{lemma:pi}
Assume that states and actions are tabular and assume that the critic is Bayes-optimal. Let $\pi'(a \mid s, s_g)$ be the goal-conditioned policy obtained after one iteration of contrastive RL with a filtering parameter of $\epsilon$. Then this policy achieves higher rewards than the initial goal-conditioned policy:
{\footnotesize \begin{equation*}
    \E_{\color{blue}\pi'(\tau \mid s_g)}\left[\sum_{t=0}^\infty \gamma^t r_{s_g}(s_t, a_t) \right] \ge \E_{\color{blue}\pi(\tau \mid s_g)}\left[\sum_{t=0}^\infty \gamma^t r_{s_g}(s_t, a_t) \right]  - \frac{2 \gamma \epsilon}{1 - \gamma} \qquad \text{for all goals} \; s_g \in \{s_g \mid p_g(s_g) > 0\}.
\end{equation*}}\!\!
\end{lemma}
The proof is in Appendix~\ref{appendix:proofs}. This result shows that performing contrastive RL on static dataset results in one step of approximate policy improvement. Re-collecting data and then applying contrastive RL over and over again corresponds to approximate policy improvement (see~\citep[Lemma 6.2]{bertsekas1996neuro}).

In summary, we have shown that applying contrastive learning to a particular choice of inputs results in an RL algorithm, one that learns a Q-function and (under some assumptions) converges to the reward-maximizing policy. Contrastive RL (NCE) is simple: it does not require multiple Q-values~\citep{fujimoto2018addressing}, target Q networks~\citep{mnih2013playing}, data augmentation~\citep{laskin2020reinforcement, kostrikov2020image}, or auxiliary objectives~\citep{srinivas2020curl, Yarats2021ImprovingSE}.

\subsection{C-learning as Contrastive Learning}
\label{sec:c-learning}

C-learning~\citep{Eysenbach2021CLearningLT} is a special case of contrastive RL: it learns a critic function to distinguish future goals from random goals. Compared with contrastive RL (NCE), C-learning learns the classifier using temporal difference learning.\footnote{The objectives are subtly different: C-learning estimates the probability that policy $\pi(\cdot \mid \cdot, s_g)$ visits state $s_f = s_g$, whereas contrastive RL (NCE) estimates the probability that any of the goal conditioned policies visit state $s_f$.} 
Viewing C-learning as a special case of contrastive RL suggests that contrastive RL algorithms might be implemented in a variety of different ways, each with relative merits. For example, \texttt{contrastive RL (NCE)} is much simpler than C-learning and tends to perform a bit better. Appendix~\ref{appendix:mc} introduces another member of the contrastive RL family (\texttt{contrastive RL (NCE + C-learning)}) that tends to yield the best performance .

\section{Experiments}
\label{sec:experiments}

Our experiments use goal-conditioned RL problems to compare contrastive RL algorithms to prior non-contrastive methods, including those that use data augmentation and auxiliary objectives. We then compare different members of the contrastive RL family, and show how contrastive RL can be effectively applied to the offline RL setting. Appendices~\ref{appendix:hparams},~\ref{appendix:experiments}, and~\ref{appendix:failed} contain experiments, visualizations, and failed experiments. %

\begin{figure}[t]
    \centering
    \vspace{-2em}
    \begin{subfigure}[b]{0.49\textwidth}
        \includegraphics[width=\linewidth]{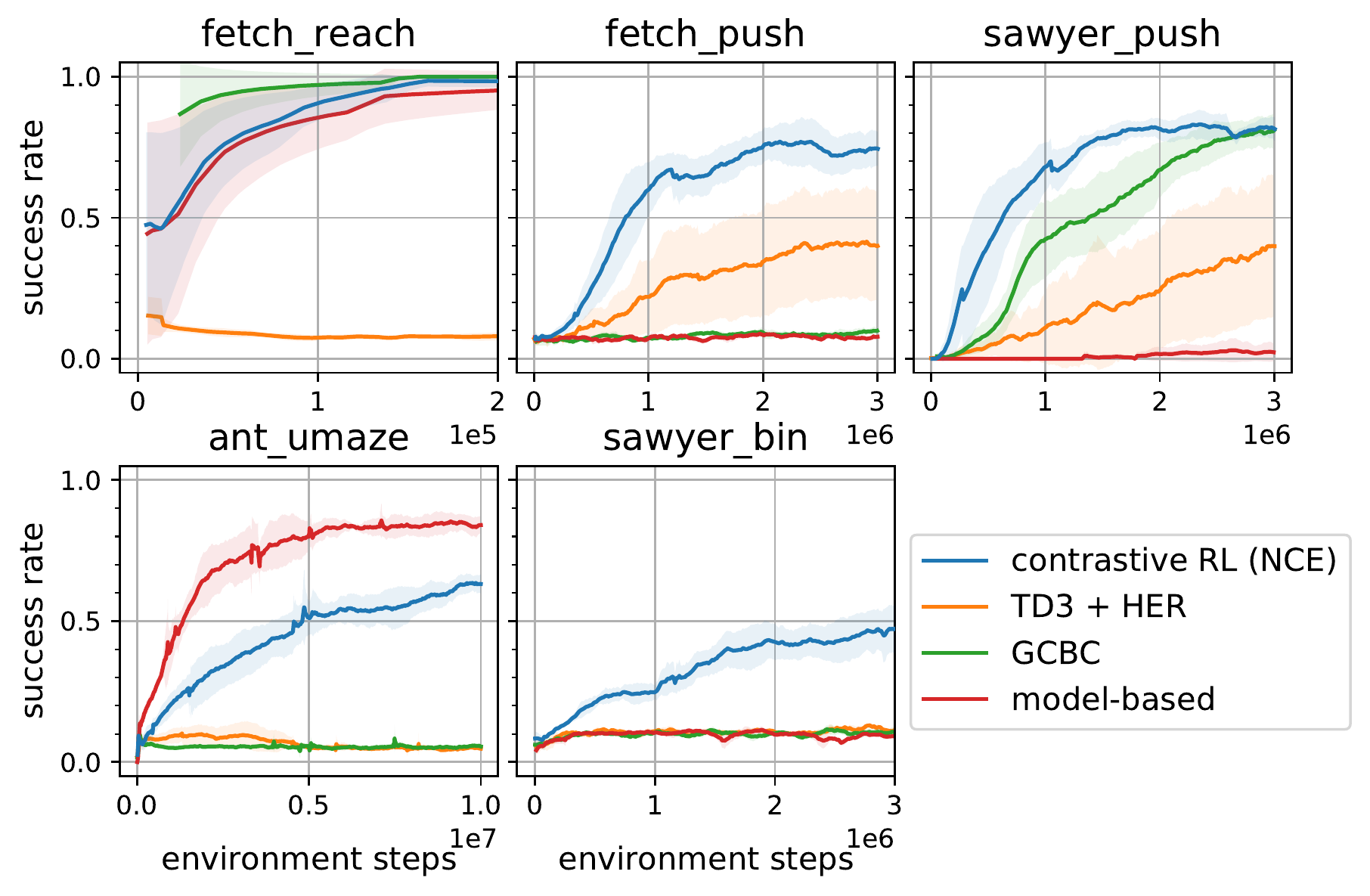}
        \caption{State-based tasks} \label{fig:benchmark-state}
    \end{subfigure}%
    ~
    \begin{subfigure}[b]{0.49\textwidth}
        \includegraphics[width=\linewidth]{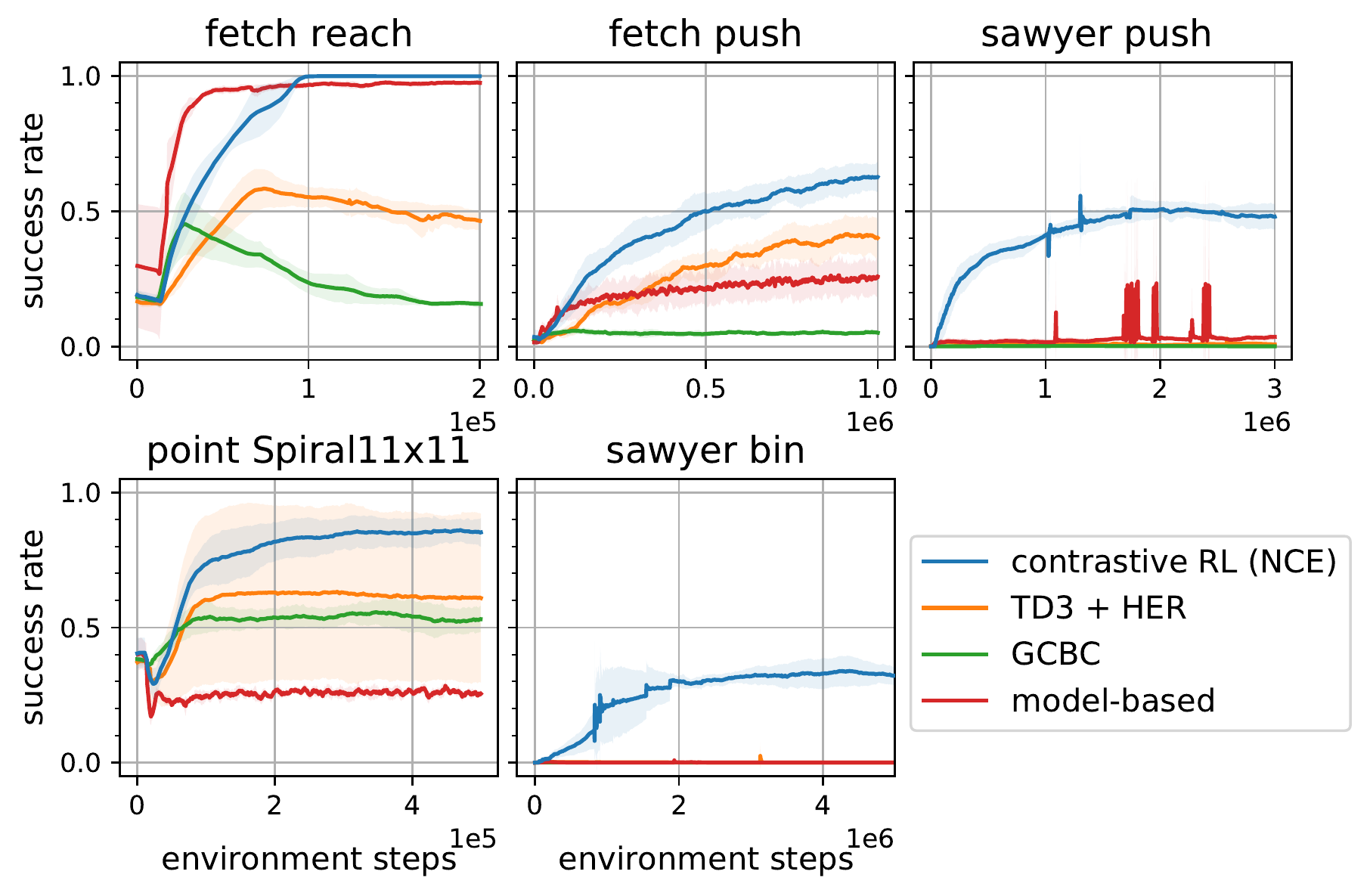}
        \caption{Image-based tasks}
        \label{fig:benchmark-img}
    \end{subfigure}
    \caption{
    \footnotesize 
    \textbf{Goal-conditioned RL.}
    Contrastive RL (NCE) outperforms prior methods on most tasks.
    \textbf{Baselines}: \texttt{HER}~\citep{Lin2019ReinforcementLW} is a prototypical actor-critic method that uses hindsight relabeling~\citep{andrychowicz2017hindsight}; \texttt{Goal-conditioned behavioral cloning (GCBC)}~\citep{ding2019goal, ghosh2020learning, srivastava2019training, lynch2020learning} performs behavior cloning on relabeled experience; \texttt{model-based} fits a density model to the discounted state occupancy measure, similar on~\citep{dayan1993improving, dosovitskiy2016learning, janner2020gamma}.
    }
    \label{fig:benchmark}
    \vspace{-1.5em}
\end{figure}

\subsection{Comparing to prior goal-conditioned RL methods}

\begin{wrapfigure}[9]{R}{0.5\textwidth}
    \vspace{-1em}
    \centering
    \begin{subfigure}[b]{0.3\linewidth}
        \centering
        \includegraphics[width=\linewidth]{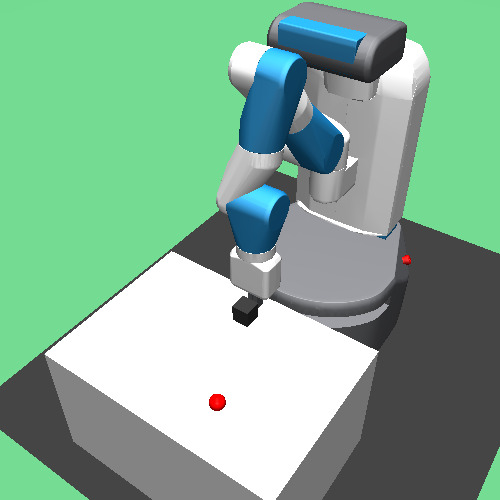}
    \end{subfigure}
    \begin{subfigure}[b]{0.3\linewidth}
        \centering
        \includegraphics[width=\linewidth]{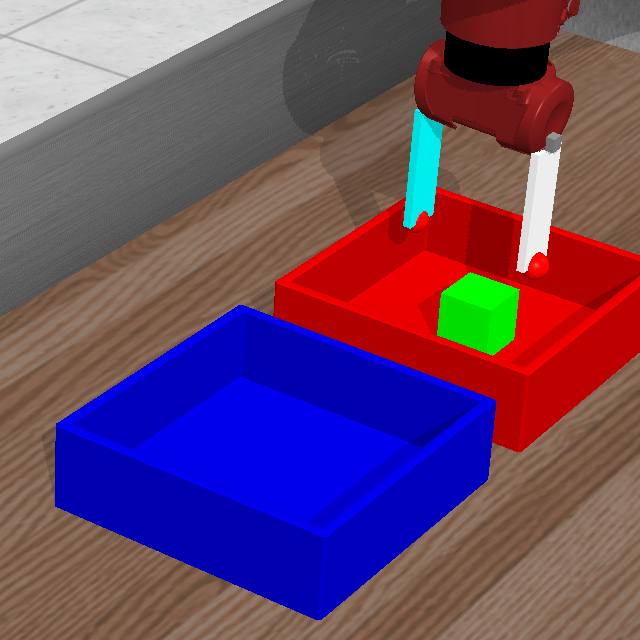}
    \end{subfigure}
    \begin{subfigure}[b]{0.3\linewidth}
        \centering
        \includegraphics[width=\linewidth]{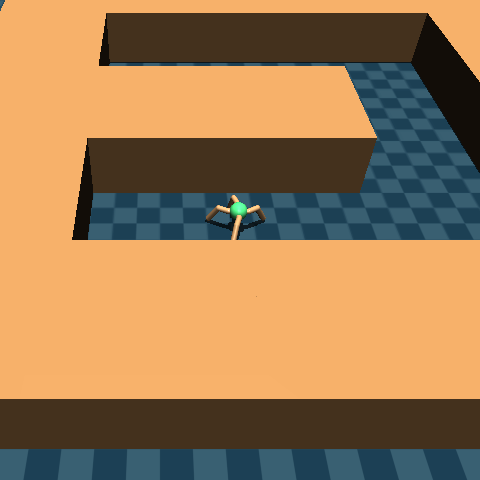}
    \end{subfigure}
    \caption{ \footnotesize \textbf{Environments.} We show a subset of the goal-conditioned environments used in our experiments.}
    \label{fig:envs}
\end{wrapfigure}
\paragraph{Baselines.} We compare three baselines. ``HER''~\citep{Lin2019ReinforcementLW} is a goal-conditioned RL method that uses hindsight relabeling~\citep{andrychowicz2017hindsight} with a high-performance actor-critic algorithm (TD3). This baseline is representative of a large class of prior work that uses hindsight relabeling~\citep{andrychowicz2017hindsight, schaul2015universal, riedmiller2018learning, levy2017learning}. Like contrastive RL, this baseline does not assume access to a reward function. The second baseline is goal-conditioned behavioral cloning (``GCBC'')~\citep{ding2019goal, ghosh2020learning, lynch2020learning, sun2019policy, emmons2021rvs, chen2021decision, srivastava2019training, paster2020planning}, which trains a policy to reach goal $s_g$ by performing behavioral cloning on trajectories that reach state $s_g$.
GCBC is a simple method that achieves excellent results~\citep{chen2021decision, emmons2021rvs} and has the same inputs as our method ($(s, a, s_f)$ triplets).
A third baseline is a model-based approach that fits a density model to the future state distribution $p^{\pi(\cdot \mid \cdot)}(s_{t+} \mid s, a)$ and trains a goal-conditioned policy to maximize the probability of the commanded goal. This baseline is similar to successor representations~\citep{dayan1993improving} and prior multi-step models~\citep{janner2020gamma, dosovitskiy2016learning}.
Both contrastive RL (Alg.~\ref{alg:contrastive}) and this model-based approach encode the future state distribution, but the output dimension of this model-based method depends on the state dimension. We, therefore, expect this approach to excel in low-dimensional settings but struggle with image-based tasks.
Where possible, we use the same hyperparameters for all methods. We will include additional representation learning baselines when studying representations in the subsequent section.

\paragraph{Tasks.} We compare it to a suite of goal-conditioned tasks, mostly taken from prior work. Four standard manipulation tasks include \texttt{fetch reach} and \texttt{fetch push} from~\citet{plappert2018multi} and \texttt{sawyer push} and \texttt{sawyer bin} from~\citet{yu2020meta}. We evaluate these tasks both with state-based observations and (unlike most prior work) image-based observations. The \texttt{sawyer bin} task poses an exploration challenge, as the agent must learn to pick up an object from one bin and place it at a goal location in another bin; the agent does not receive any reward shaping or demonstrations. We include two navigation tasks: \texttt{point Spiral11x11} is a 2D maze task with image observations and \texttt{ant umaze}~\citep{fu2020d4rl} is a 111-dimensional locomotion task that presents a challenging low-level control problem.
Where possible, we use the same initial state distribution, goal distribution, observations, and definition of success as prior work. Goals have the same dimension as the states, with one exception: on the \texttt{ant umaze} task, we used the global $XY$ position as the goal. We illustrate three of the tasks to the right. The agent does not have access to any ground truth reward function.

We report results in Fig.~\ref{fig:benchmark}, using five random seeds for each experiment and plotting the mean and standard deviation across those random seeds.
On the state-based tasks (Fig.~\ref{fig:benchmark-state}), most methods solve the easiest task (\texttt{fetch reach}) while only our method solves the most challenging task (\texttt{sawyer bin}). Our method also outperforms all prior methods on the two pushing tasks. The model-based baseline performs best on the \texttt{ant umaze} task, likely because learning a model is relatively easy when the goal is lower-dimensional (just the $XY$ location). On the image-based tasks (Fig.~\ref{fig:benchmark-img}), most methods make progress on the two easiest tasks (\texttt{fetch reach} and \texttt{point Spiral11x11}); our method outperforms the baselines on the three more challenging tasks. Of particular note is the success on \texttt{sawyer push} and \texttt{sawyer bin}: while the success rate of our method remains below 50\%, no baselines make any progress on learning these tasks.
These results suggest that contrastive RL (NCE) is a competitive goal-conditioned RL algorithm.

\begin{figure}[t]
    \centering
    \vspace{-2em}
    \includegraphics[width=\linewidth]{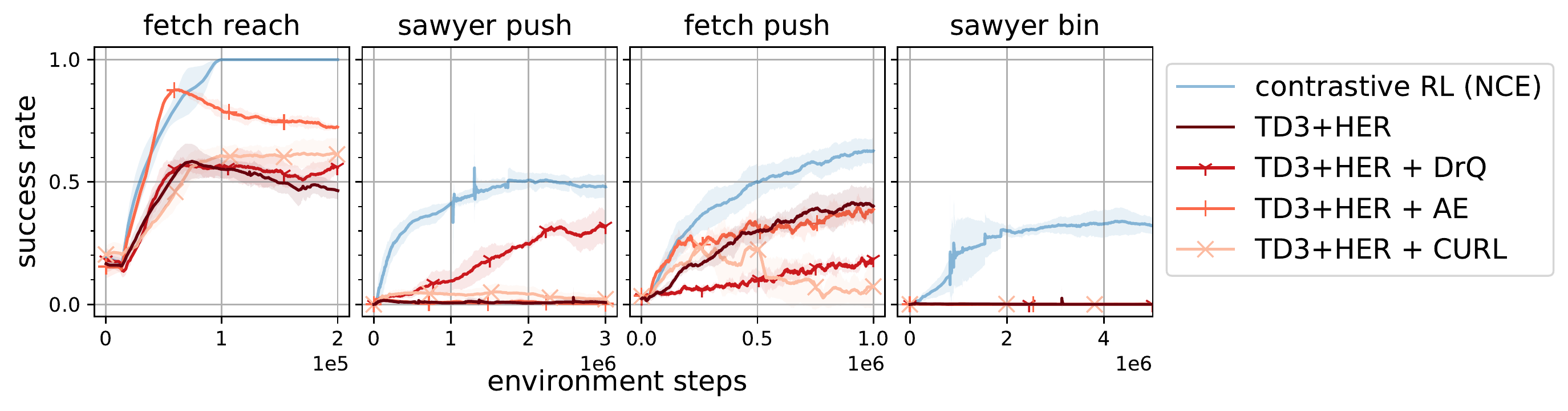}
    \vspace{-1.5em}
    \caption{
    \footnotesize 
    \textbf{Representation learning for image-based tasks.}
    While adding data augmentation and auxiliary representation objectives can boost the performance of the \texttt{TD3+HER} baseline,
    replacing the underlying goal-conditioned RL algorithm with one that resembles contrastive representation learning (i.e., ours) yields a larger increase in success rates.
    \textbf{Baselines}: \texttt{DrQ}~\citep{kostrikov2020image} augments images and averages the Q-values across 4 augmentations; \texttt{auto encoder (AE)} adds an auxiliary reconstruction loss~\citep{finn2016deep, nair2018visual, Nasiriany2019PlanningWG, Yarats2021ImprovingSE}; \texttt{CURL}~\citep{srinivas2020curl} applies RL on top of representations learned via augmentation-based contrastive learning.}
    \label{fig:her}
    \vspace{-1.5em}
\end{figure}

\subsection{Comparing to prior representation learning methods}

We hypothesize that contrastive RL may automatically learn good representations. To test this hypothesis, we compare contrastive RL (NCE) to techniques proposed by prior work for representation learning. These include data augmentation~\citep{kostrikov2020image,yarats2021mastering, laskin2020reinforcement} (``DrQ'') and auxiliary objectives based on an autoencoder~\citep{finn2016deep, nair2018visual, Nasiriany2019PlanningWG, Yarats2021ImprovingSE} (``AE'') and a contrastive learning objective (``CURL'') that generates positive examples using data augmentation, similar to prior work~\citep{nachum2018near,srinivas2020curl,stooke2021decoupling}. Because prior work has demonstrated these techniques in combination with actor-critic RL algorithms, we will use these techniques in combination with the actor-critic baseline from the previous section (``TD3 + HER''). While contrastive RL (NCE) resembles a contrastive representation learning method, it does not include any data augmentation or auxiliary representation learning objectives.

We show results in Fig.~\ref{fig:her}, with error bars again showing the mean and standard deviation across 5 random seeds.
While adding the autoencoder improves the baseline on the \texttt{fetch reach} and adding DrQ improves the baseline on the \texttt{sawyer push}, contrastive RL (NCE) outperforms the prior methods on all tasks. Unlike these methods, contrastive RL does not use auxiliary objectives or additional domain knowledge in the form of image-appropriate data augmentations. These experiments do not show that representation learning is never useful, and do not show that contrastive RL cannot be improved with additional representation learning machinery. Rather, they show that designing RL algorithms that structurally resemble contrastive representation learning yields bigger improvements than simply adding representation learning tricks on top of existing RL algorithms.

\begin{figure}[t]
    \centering
    \vspace{-2em}
    \begin{subfigure}[b]{0.49\textwidth}
        \centering
        \includegraphics[width=\linewidth]{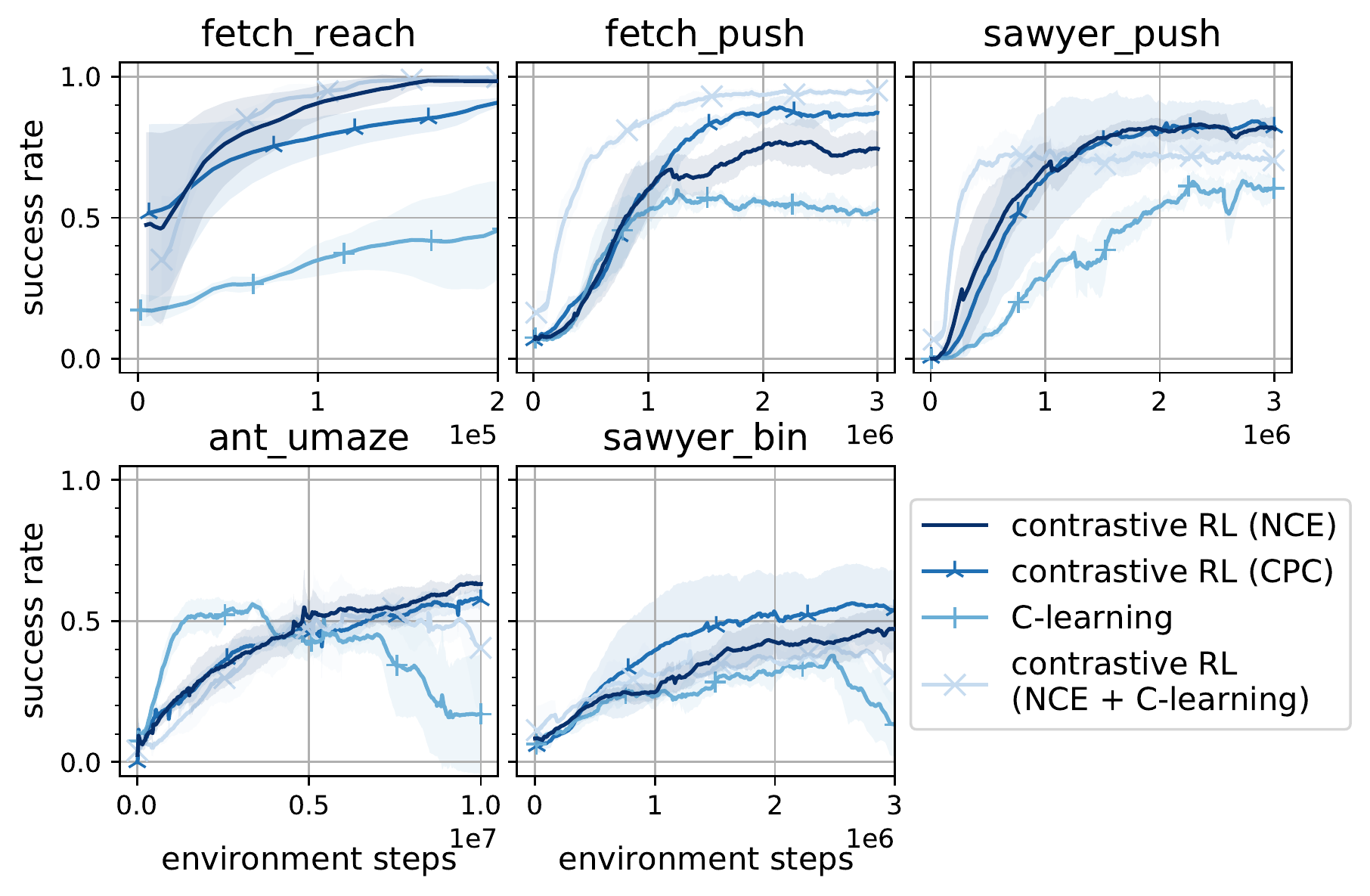}
        \vspace{-0.5em}
        \caption{state-based observations}
    \end{subfigure}%
    ~
    \begin{subfigure}[b]{0.49\textwidth}
        \centering
        \includegraphics[width=\linewidth]{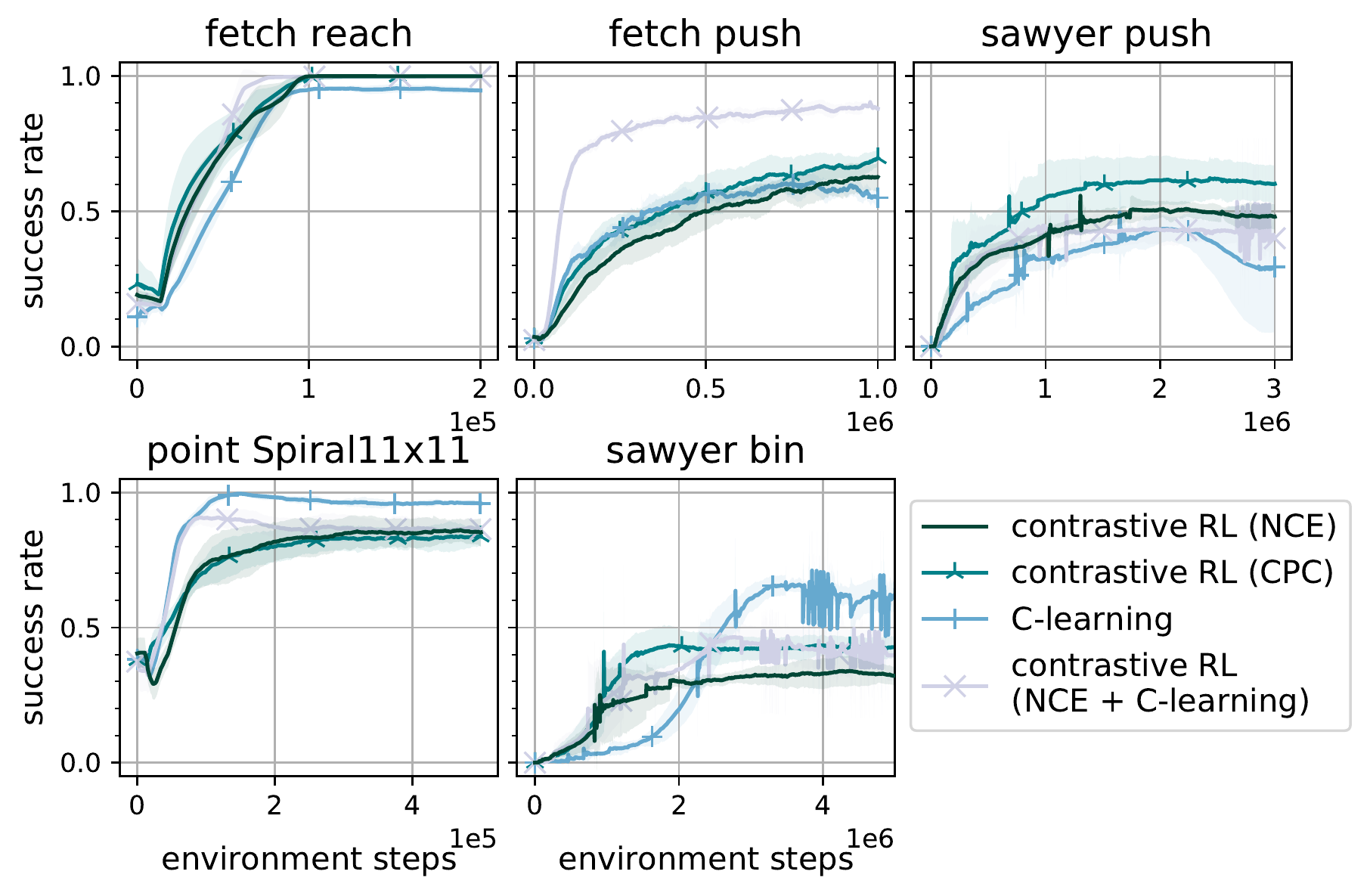}
        \vspace{-0.5em}
        \caption{image-based observations}
    \end{subfigure}
    \vspace{-0.5em}
    \caption{
    \footnotesize
    \textbf{Contrastive RL design decisions.}
    Generalizing C-learning to a family of contrastive RL algorithms allowed us to identify algorithms that are much simpler (contrastive RL (NCE)) and that consistently achieve higher performance (contrastive RL (NCE + C-learning)).\label{fig:benchmark-ours}}
    \vspace{-1em}
\end{figure}

\subsection{Probing the dimensions of contrastive RL}
\label{sec:benchmark-ours}

Up to now, we have focused on the specific instantiation of contrastive RL spelled out in Alg.~\ref{alg:contrastive}. However, there is a whole family of RL algorithms with contrastive characteristics.
C-learning is a contrastive RL algorithm that uses temporal difference learning (Sec.~\ref{sec:c-learning}). Contrastive RL (CPC) is a variant of Alg.~\ref{alg:contrastive} based on the infoNCE objective~\citep{oord2018representation} that we derive in Appendix~\ref{appendix:cpc}
Contrastive RL (NCE + C-learning) is a variant that combines C-learning with Alg.~\ref{appendix:mc} (see Appendix~\ref{appendix:mc}.).
The aim of these experiments are to study whether generalizing C-learning to a family of contrastive RL algorithms was useful: do the simpler methods achieve similar performance, and do other methods achieve better performance?

We present results in Fig.~\ref{fig:benchmark-ours}, again plotting the mean and standard deviation across five random seeds. Contrastive RL (CPC) outperforms contrastive RL (NCE) on three, suggesting that swapping one mutual information estimator for another can sometimes improve performance, though both estimators can be effective.
C-learning outperforms contrastive RL (NCE) on three tasks but performs worse on other tasks. Contrastive RL (NCE + C-learning) consistently ranks among the best methods.  These experiments demonstrate that the prior contrastive RL method, C-learning~\citep{Eysenbach2021CLearningLT}, achieves good results on most tasks; generalizing  C-learning to a family of contrastive RL algorithms resulting in  new algorithms that achieve higher performance and can be much simpler.

\subsection{Partial Observability and Moving Cameras}

\begin{wrapfigure}[10]{R}{0.4\textwidth}
    \vspace{-2em}
    \includegraphics[width=\linewidth]{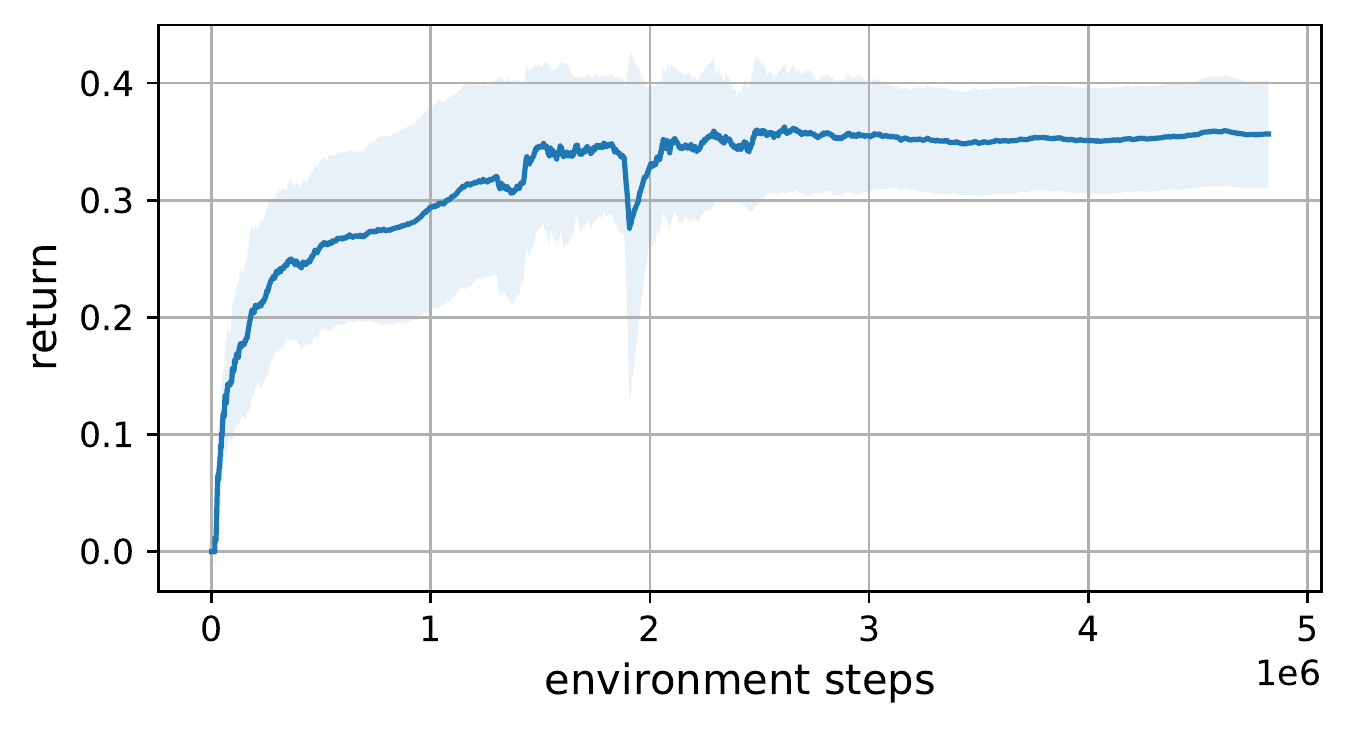}
    \vspace{-2em}
    \caption{
    \footnotesize\textbf{Partial observability and moving cameras.} Contrastive RL can solve partially observed tasks.}
    \label{fig:partial}
\end{wrapfigure}

Many realistic robotics tasks exhibit partial observability, and have cameras that are not fixed but rather attached to moving robot parts. Our next experiment tests if contrastive RL can cope with these sorts of challenges. To study this question, we modified the \texttt{sawyer push} task so that the camera tracks the hand at a fixed distance, as if it were rigidly mounted to the arm. This means that, at the start of the episode, the scene is occluded by the wall at the edge of the table, so the agent cannot see the location of the puck (see Fig.~\ref{fig:partial} (left)). Nonetheless, contrastive RL (NCE) successfully handles this partial observability, achieving a success rate of around 35\%. Fig.~\ref{fig:partial} (left) shows an example rollout and Fig.~\ref{fig:partial} (right) shows the learning curve. For comparison, the success rate when using the fixed static camera was ~75\%. Taken together, these results suggest that contrastive RL can cope with moving cameras and partial observability, while also suggesting that improved strategies (e.g., non-Markovian architectures) might achieve even better results.

\subsection{Contrastive RL for Offline RL}
\label{sec:offline}

Our final experiment studies whether the benefits from contrastive RL (NCE) transfer to the offline RL setting, where the agent is prohibited from interacting with the environment. We use the benchmark AntMaze tasks from the D4RL benchmark~\citep{fu2020d4rl}, as these are goal-conditioned tasks commonly studied in the offline setting. 

We adapt contrastive RL (NCE) to the offline setting by adding an additional (goal-conditioned) behavioral cloning term to the policy objective (Eq.~\ref{eq:actor-loss}), using a coefficient of $\lambda$:
{\footnotesize \begin{equation*}
    \max_{\pi(a \mid s, s_g)} \E_{\pi(a \mid s, s_g)p(s, a_\text{orig}, s_g)}\left[(1 - \lambda) \cdot f(s, a, s_f = s_g) + \lambda \cdot \log \pi(a_\text{orig} \mid s, s_g) \right].
\end{equation*}}\!\!
Note that setting $\lambda = 1$ corresponds to GCBC~\citep{ding2019goal, ghosh2020learning, lynch2020learning, sun2019policy, emmons2021rvs, chen2021decision, srivastava2019training, paster2020planning}, which we will include as a baseline. 
Following TD3+BC~\citep{fujimoto2021minimalist}, we learn multiple critic functions (2 and 5) and take the minimum when computing the actor update. We also compare to prior offline RL methods that eschew TD learning: (unconditional) behavioral cloning (BC), the implementation of GCBC from~\citep{emmons2021rvs} (which refers to GCBC as RvS-G), and a recent method based on the transformer architecture (DT~\citep{chen2021decision}). Lastly, we compare with two more complex methods that use TD learning: TD3+BC~\citep{fujimoto2021minimalist} and IQL~\citep{kostrikov2021offline}. Unlike contrastive RL and GCBC, these TD learning methods do not perform goal relabeling. We use the numbers reported for these baselines in prior work~\citep{emmons2021rvs, kostrikov2021offline}.

\begin{table}[t]
    \centering
    \vspace{-2em}
    \caption{\footnotesize \textbf{Offline RL on D4RL AntMaze}~\citep{fu2020d4rl}. Contrastive RL outperforms all baselines in 5 out of 6 tasks. 
    }
    \label{tab:offline-rl}
    \footnotesize 
    \begin{tabular}{c|ccccc|cc}
        \toprule
        \multicolumn{1}{c}{} & \multicolumn{5}{c}{no TD} & \multicolumn{2}{c}{uses TD}\\ \cmidrule(r{4pt}){2-6} \cmidrule(l){7-8}
         & BC & DT & GCBC & \multicolumn{2}{c|}{Contrastive RL + BC} & TD3+BC$^*$ & IQL$^*$ \\
        &&&& 2 nets & 5 nets \\
        \midrule
        umaze-v2 &  54.6 & 65.6 & 65.4 & 81.9 {(\tiny $\pm$1.7)} & 79.8 {(\tiny $\pm$1.4)} & 78.6 & {\bf87.5} \\
        umaze-diverse-v2 & 45.6 & 51.2 & 60.9 & {\bf 75.4 {(\tiny $\pm$3.5)}} & {\bf 77.6 {(\tiny $\pm$2.8)}} & 71.4 & 62.2 \\
        medium-play-v2 & 0.0 & 1.0 & 58.1 & {\bf 71.5 {(\tiny $\pm$5.2)}} & {\bf 72.6 {(\tiny $\pm$2.9)}} & 10.6 & {\bf 71.2} \\
        medium-diverse-v2 & 0.0 & 0.6 & 67.3 & {\bf 72.5 {(\tiny $\pm$2.8)}} & {\bf 71.5 {(\tiny $\pm$1.3)}} & 3.0 & {\bf 70.0} \\
        large-play-v2 & 0.0 & 0.0 & 32.4 & 41.6 {(\tiny $\pm$6.0)} & {\bf 48.6 {(\tiny $\pm$4.4)}} & 0.2 & 39.6 \\
        large-diverse-v2 & 0.0 & 0.2 & 36.9 & {\bf 49.3 {(\tiny $\pm$6.3)}} & {\bf54.1 {(\tiny $\pm$5.5)}} & 0.0 & 47.5 \\
        \multicolumn{8}{l}{\scriptsize \qquad $^*$ While TD3+BC and IQL report results on the \texttt{-v0} tasks, the change to \texttt{-v2} has a negligible effect on TD methods~\citep{iql2022}.} \\ \bottomrule
    \end{tabular}
    \vspace{-1em}
\end{table}

As shown in Table~\ref{tab:offline-rl}, contrastive RL (NCE) outperforms all baselines on five of the six benchmark tasks. Of particular note are the most challenging ``-large'' tasks, where contrastive RL achieves a $7\%$ to $9\%$ absolute improvement over IQL. We note that IQL does not use goal relabeling, which is the bedrock of contrastive RL. Compared to baselines that do not use TD learning, the benefits are more pronounced, with a median (absolute) improvement over GCBC of 15\%. The performance of contrastive RL improves when increasing the number of critics from $2$ to $5$, suggesting that the key to solving more challenging offline RL tasks may be increased capacity, rather than TD learning. Taken together, these results show the value of contrastive RL for offline goal-conditioned tasks.

\section{Conclusion}
\label{sec:conclusion}

In this paper, we showed how contrastive representation learning can be used for goal-conditioned RL. This connection not only lets us re-interpret a prior RL method as performing contrastive learning, but also suggests a family of contrastive RL methods, which includes simpler algorithms, as well as algorithms that attain better overall performance. While this paper might be construed to imply that RL is more or less important than representation learning~\citep{langford_2010, lecun2016predictive, silver2021reward, srinivas2021unsupervised}, we have a different takeaway: that it may be enough to build RL algorithms that \emph{look like} representation learning.

One limitation of this work is that it looks only at the goal-conditioned RL problems. How these methods might be applied to arbitrary RL problems remains an open problem, though we note that recent algorithms for this setting~\citep{eysenbach2021replacing} already bear a resemblance to contrastive RL. Whether the rich set of ideas from contrastive learning might be used to construct even better RL algorithms likewise remains an open question.

{\footnotesize

\paragraph{Acknowledgements.}
Thanks to Hubert Tsai, Martin Ma, and Simon Kornblith for discussions about contrastive learning. Thanks to Kamyar Ghasemipour, Suraj Nair, and anonymous reviewers for feedback on the paper. Thanks to Ofir Nachum, Daniel Zheng, and the JAX and Acme teams for helping to release and debug the code. This material is supported by the Fannie and John Hertz Foundation and the NSF GRFP (DGE1745016). UC Berkeley research is also supported by gifts from Alibaba, Amazon Web Services, Ant Financial, CapitalOne, Ericsson, Facebook, Futurewei, Google, Intel, Microsoft, Nvidia, Scotiabank, Splunk and VMware.

}
\clearpage
\appendix

\section{Additional Related Work}
\label{appendix:related}

\begin{wrapfigure}{R}{0.5\textwidth}
    \centering
    \includegraphics[width=\linewidth]{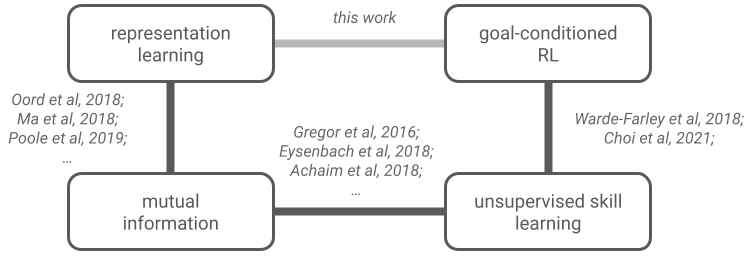}
    \caption{\footnotesize \textbf{Connecting related work.} This work helps draw connections between prior work, filling in a missing link.}
    \label{fig:related}
\end{wrapfigure}
Our work is also related to unsupervised skill discovery~\citep{gregor2016variational, eysenbach2018diversity,achiam2018variational,sharma2019dynamics, hansen2019fast,laskin2021cic, choi2021variational}, in that the algorithm learns multiple policies by interacting in the environment without a reward function. Both these skill learning algorithms and our contrastive algorithm optimize a lower bound on mutual information. Indeed, prior work has discussed the close connection between mutual information and goal-conditioned RL~\citep{choi2021variational, warde2018unsupervised}. 
The key challenge in making this connection is \emph{grounding} the skills, so that each skill corresponds to a specific goal-conditioned policy. While the skills can be grounded by manually-specifying the critic used for maximizing mutual information~\citep{choi2021variational}, manually-specifying the critic for high-dimensional tasks (e.g., images) would be challenging. Our work takes a different approach to grounding, one based on reasoning directly about continuous probabilities. In the end, our method will learn skills that each corresponds to a specific goal-conditioned policy and will be scalable to high-dimensional tasks.

Fig.~\ref{fig:related} highlights some of the connections between related work. Prior work has thoroughly explained how many representation learning methods correspond to a lower bound on mutual information~\citep{poole2019variational, ma2018noise}. Prior work in RL has proposed unsupervised skill learning algorithms using similar mutual information objectives~\cite{achiam2018variational, eysenbach2018diversity, gregor2016variational}, and more recent work has connected these unsupervised skills learning algorithms to goal-reaching. The key contribution of this paper is to connect representation learning to goal-conditioned RL.

\section{Proofs}
\label{appendix:proofs}

\subsection{Q-function are equivalent to the discounted state occupancy measure}
This section proves Proposition~\ref{prop:q}. We start by recalling the definition of the discounted state occupancy measure (Eq.~\ref{eq:future-states}):
\begin{align}
    p(s_{t+} = s_g) = (1 - \gamma) \sum_{t=0}^\infty \gamma^t p_t^{\pi(\cdot \mid \cdot, s_g)}(s_t = s_g). \label{eq:future-2}
\end{align}
We first analyze the term for $t = 0$, and then analyze the term for $t > 0$. The probability of visiting a state at time $t = 0$ is just the initial state distribution:
\begin{align*}
    p_0^{\pi(\cdot \mid \cdot, s_g)}(s_t = s_g) = p_0(s_0 = s_g).
\end{align*}
We can now rewrite Eq.~\ref{eq:future-2} as
\begin{align}
    p(s_{t+} = s_g) = (1 - \gamma) p_0(s_0 = s_g) + (1 - \gamma) \sum_{t=1}^\infty \gamma^t p_t^{\pi(\cdot \mid \cdot, s_g)}(s_t = s_g). \label{eq:future-3}
\end{align}
For $t > 1$, we can write the term as follows:
\begin{align*}
    p_t^{\pi(\cdot \mid \cdot, s_g)}(s_t = s_g) 
    & = \E_{p_{t-1}^{\pi(\cdot \mid \cdot, s_g)}(s_{t-1})\pi(a_{t-1} \mid s_{t-1}, s_g)}\left[ p_t(s_t = s_g \mid s_{t-1}, a_{t-1})\right] \\
    & = \E_{p_{t-1}^{\pi(\cdot \mid \cdot, s_g)}(s_{t-1}),\pi(a_{t-1} \mid s_{t-1}, s_g)}\left[ p(s_t = s_g \mid s_{t-1}, a_{t-1}) \right] \\
    & = \E_{\tau \sim \pi(\tau \mid s_t)}\left[ p(s_t = s_g \mid s_{t-1}, a_{t-1}) \right].
\end{align*}
In the second line, we have used the Markov property to say that the probability of visiting $s_g$ at time $t$ depends only on dynamics, $p(s_{t+1} \mid s_t, a_t)$.
In the third line, we have rewritten the expectation over trajectories, using $s_{t-1}$ and $a_{t-1}$ and the $t-1^{\text{th}}$ state-action pair in the trajectory.
Substituting this into Eq.~\ref{eq:future-3}, we get
\begin{align*}
    p(s_{t+} = s_g)
    &= (1 - \gamma) p_0(s_0 = s_g) + (1 - \gamma) \sum_{t=1}^\infty \gamma^t \E_{\tau \sim \pi(\tau \mid s_g)}\left[p(s_t = s_g \mid s_{t-1}, a_{t-1}) \right] \\
    &= (1 - \gamma) p_0(s_0 = s_g) + (1 - \gamma) \sum_{t=0}^\infty \gamma^t \E_{\tau \sim \pi(\tau \mid s_g)}\left[p(s_{t+1} = s_g \mid s_{t}, a_{t}) \right] \\
    &= (1 - \gamma) p_0(s_0 = s_g) + (1 - \gamma)  \E_{\tau \sim \pi(\tau \mid s_g)}\left[ \sum_{t=0}^\infty \gamma^t p(s_{t+1} = s_g \mid s_{t}, a_{t}) \right] \\
    &= \E_{\tau \sim \pi(\tau \mid s_g)}\left[(1 - \gamma) p_0(s_0 = s_g) + (1 - \gamma)   \sum_{t=0}^\infty \gamma^t p(s_{t+1} = s_g \mid s_{t}, a_{t}) \right] \\
    &= \E_{\tau \sim \pi(\tau \mid s_g)}\left[\sum_{t=0}^\infty \gamma^t r_g(s_t, a_t) \right].
\end{align*}
On the second line, we have changed the bounds of the summation to start at 0, and changed the terms inside the summation accordingly.
On the third line, we applied linearity of expectation to move the summation inside the expectation.
On the fourth line, we applied linearity of expectation again to move the term for $t = 0$ inside the expectation.
Finally, we substituted the definition of $r_g(s, a)$ to obtain the desired result.

\subsection{Contrastive RL is Policy Improvement}
This section proves the Contrastive RL (NCE) corresponds to policy improvement, yielding policies with higher rewards at each iteration (Lemma~\ref{lemma:pi}).

\begin{proof}
The main idea of the proof is to relate the Q-values for the average policy to the Q-values for the goal-conditioned policy. We do this by employing the result from ~\citep[Appendix C.2]{eysenbach2022imitating}, where $\epsilon$ is the parameter for filtered relabeling (Sec.~\ref{sec:analysis}):
{\footnotesize \begin{align*}
\left|Q^{\beta(\cdot mid \cdot, e)}(s, a, e) - Q^{\beta(\cdot \mid \cdot, e')}(s, a, e) \right| \le \epsilon.
\end{align*}}\!\!
This result means that we are doing policy improvement with approximate Q-values. Then, \citep[Lemma 6.1]{bertsekas1996neuro} tells that doing policy improvement using approximate Q-values gives us approximate policy improvement:
{\footnotesize \begin{align*}
    \E_{\color{blue}\pi'(\tau \mid s_g)}\left[\sum_{t=0}^\infty \gamma^t r_{s_g}(s_t, a_t) \right] \ge \E_{\color{blue}\pi(\tau \mid s_g)}\left[\sum_{t=0}^\infty \gamma^t r_{s_g}(s_t, a_t) \right]  - \frac{2 \gamma \epsilon}{1 - \gamma} \qquad \text{for all goals} \; s_g \in \{s_g \mid p_g(s_g) > 0\}.
\end{align*}}\!\!
\end{proof}

\section{Contrastive RL (CPC)}
\label{appendix:cpc}

In this section, we derive a version of contrastive RL based on the infoNCE objective~\citep{oord2018representation}. Compared with the NCE objective used in contrastive RL (NCE), this objective uses a categorical cross entropy loss instead of a binary cross entropy loss. We replace Eq.~\ref{eq:nce-rl} with the following infoNCE objective~\citep{oord2018representation}:
\begin{equation*}
   \max_f \E_{\substack{(s, a) \sim p(s, a), {\color{green}s_f^{(1)}} \sim p^{\pi(\cdot \mid \cdot)}(s_{t+} \mid s, a)\\{\color{red}s_f^{(2:B)}} \sim p(s_f)}}\bigg[ \log p^{(1)} \bigg],
\end{equation*}
where $p^{(1)}$ is the first coordinate of the softmax over the critic:
\begin{equation*}
    p = \textsc{SoftMax}([f(s, a, s_f^{(1)}), \cdots, f(s, a, s_f^{(b)})]).
\end{equation*}
The optimal critic for the infoNCE loss satisfies~\citep{oord2018representation, poole2019variational, ma2018noise}
\begin{equation*}
    f^*(s, a, s_f) = \log \left( \frac{p^{\pi(\cdot \mid \cdot)}(s_{t+} = s_f \mid s, a)}{p(s_f)c(s, a)} \right),
\end{equation*}
where $c(s, a)$ is an arbitrary function.
Thus, there are many optimal critics. Choosing actions that maximize the critic $f^*$ does not necessarily correspond to choosing actions that maximize the probability of the future state. Thus, we need to regularize $c(s, a)$ so that it does not depend on $a$. We do this by introducing a regularizer, based on~\citep{tsai2020neural}:
\begin{equation*}
    \min_f E_{s_f^{(1:B)} \sim p(s_f)}\textsc{LogSumExp}([f(s, a, s_f^{(1)}), \cdots, f(s, a, s_f^{(b)})])^2.
\end{equation*}

To provide some intuition for this regularizer, consider applying this regularizer to an optimal critic:
\begin{align*}
    & \textsc{LogSumExp}([f^*(s, a, s_f^{(1)}), \cdots, f^*(s, a, s_f^{(b)})])^2 \\
    & = \left( \log \frac{1}{c(s, a)} \sum_{s_f} \frac{p^{\pi(\cdot \mid \cdot)}(s_{t+} = s_f \mid s, a)}{p(s_f)c(s, a)} \right)^2 \\
    & = \left( \log \sum_{s_f \in s_f^{(1:B)}} \frac{p^{\pi(\cdot \mid \cdot)}(s_{t+} = s_f \mid s, a)}{p(s_f)}  - \log c(s, a) \right)^2 \\
    & \approx \left( \log \sum_{s_f \in s_f^{(2:B)}} \frac{p^{\pi(\cdot \mid \cdot)}(s_{t+} = s_f \mid s, a)}{p(s_f)}  - \log c(s, a)\right)^2 \\
    & \approx \left( \log \E_{s_f \sim p(s_f)}\left[\frac{p^{\pi(\cdot \mid \cdot)}(s_{t+} = s_f \mid s, a)}{p(s_f)} \right]  - \log c(s, a)\right)^2 \\
    & = \left( - \log c(s, a)\right)^2.
\end{align*}
In the third line we ignore the positive term; this is reasonable if the batch size is large enough. In the third line we replaced the sum with an expectation; this is biased because $\log(\cdot)$ is not a linear function.
Thus, this regularizer (approximately) regularizes $c(s, a)$ to be close to 1 for all states and actions. By reducing the dependency of $c(s, a)$ on the actions $a$, we can ensure that actions that maximize the critic do maximize the probability of reaching the desired goal. In practice, we add this regularizer with the infoNCE objective, using a coefficient of 1e-2 on the regularizer.

\section{Contrastive RL (NCE + C-learning)}
\label{appendix:mc}

In this section we describe \texttt{contrastive RL (NCE + C-learning)} the combined NCE + C-learning method used in Sec.~\ref{sec:benchmark-ours} (Fig.~\ref{fig:benchmark-ours}). Mathematically, the NCE + C-learning objective is a simple, unweighted sum of the {\color{orange}C-learning objective} and the {\color{blue}NCE objective}:
{\footnotesize \begin{align*}
    \gL(f) = & \color{orange} (1 - \gamma) \E_{(s, a) \sim p(s, a), s_f^+ \sim p(s_{t+1} \mid s_t, a_t)}[\log \sigma(f(s, a, s_f^+))] \\
    & \color{orange} + \gamma \E_{\substack{ s_g \sim p_g(s_g), (s_t, a_t) \sim p(s, a), \\s_{t+1} \sim p(s_{t+1} \mid s_t, a_t), a_{t+1} \sim \pi(a_{t+1} \mid s_{t+1}, s_g)}} \bigg[\underbrace{\frac{p(s_{t+} = s_g \mid s_t, a_t)}{p(s_f = s_g)}}_{\approx \exp(f(s_{t+1}, a_{t+1}, s_g))} \log \sigma(f(s, a, s_f = s_g)) \bigg] \\
    & \color{orange} + \E_{s_g \sim p_g(s_g), (s, a) \sim p(s, a)} \left[\log (1 - \sigma(f(s, a, s_g)) \right] \\
    & \color{blue} + \E_{(s, a) \sim p(s, a), s_f^+ \sim p(s_{t+} \mid s_t, a_t)}[\log \sigma(f(s, a, s_f^+))] + \E_{(s, a) \sim p(s, a), s_f^- \sim p(s_f)}[\log (1 - \sigma(f(s, a, s_f^-)))].
\end{align*}}\!\!
While we could use half the batch to compute each of the loss terms, we can increase the effective sample size by being careful with how the terms are estimated. First, we note that the first two terms of each loss are similar -- sample a future state (either the next state or a future state) and label it as a positive. We can thus combine these two terms by sampling from a mixture of these two distributions,
\begin{equation*}
    \tilde{p}(s_f \mid s_t, a_t) = \frac{1 - \gamma}{1 + 1 - \gamma} p(s_{t+1} = s_f \mid s_t, a_t) + \frac{1}{1 + 1 - \gamma} p(s_{t+} = s_f \mid s_t, a_t),
\end{equation*}
and scaling the resulting loss by $1 + 1 - \gamma = 2 - \gamma$:
{\footnotesize \begin{align*}
    \gL_1(f) \triangleq & \color{orange} (1 - \gamma) \E_{(s, a) \sim p(s, a), s_f^+ \sim p(s_{t+1} \mid s_t, a_t)}[\log \sigma(f(s, a, s_f^+))] + \color{blue} + \E_{(s, a) \sim p(s, a), s_f^+ \sim p(s_{t+} \mid s_t, a_t)}[\log \sigma(f(s, a, s_f^+))] \\
    &= (2 - \gamma) \E_{(s, a) \sim p(s, a), s_f^+ \sim \tilde{p}(s_f \mid s_t, a_t)}[\log \sigma(f(s, a, s_f^+))]
\end{align*}}\!\!
This trick increases the effective sample size by 96\% ($130 \rightarrow 256$, as measured using~\citep{kish1965survey}).

Both losses also contain terms that are an expectation over random goals. We can likewise combine those terms:
{\footnotesize \begin{align*}
    \gL_2(f) \triangleq & \color{orange} \gamma \E_{\substack{ s_g \sim p_g(s_g), (s_t, a_t) \sim p(s, a),\\s_{t+1} \sim p(s_{t+1} \mid s_t, a_t), a_{t+1} \sim \pi(a_{t+1} \mid s_{t+1}, s_g)}} \bigg[\lfloor \exp(f(s_{t+1}, a_{t+1}, s_g)) \rfloor_\text{sg} \log \sigma(f(s, a, s_f = s_g)) \bigg] \\
    & \color{orange} + \E_{s_g \sim p_g(s_g), (s, a) \sim p(s, a)} \left[\log (1 - \sigma(f(s, a, s_g)) \right] \color{blue} + \E_{(s, a) \sim p(s, a), s_f^- \sim p(s_f)}[\log (1 - \sigma(f(s, a, s_f^-)))] \\
    & = \gamma \E_{\substack{ s_g \sim p_g(s_g), (s_t, a_t) \sim p(s, a),\\s_{t+1} \sim p(s_{t+1} \mid s_t, a_t), a_{t+1} \sim \pi(a_{t+1} \mid s_{t+1}, s_g)}} \bigg[\lfloor \exp(f(s_{t+1}, a_{t+1}, s_g)) \rfloor_\text{sg} \log \sigma(f(s, a, s_f = s_g)) \bigg] \\
    & + 2\E_{s_g \sim p_g(s_g), (s, a) \sim p(s, a)} \left[\log (1 - \sigma(f(s, a, s_g)) \right].
\end{align*}}\!\!
Note that estimating the first term in $\gL_2$ requires sampling an action for each next state and goal pair. This prohibits us from using the same outer product trick as in Sec.~\ref{sec:complete} to estimate this term. While we could still use that trick to estimate the second term in $\gL_2$, we found that doing so hurt performance. We hypothesize that the reason is that this creates an imbalance in the gradients -- some goals are labeled as negatives but are not also labeled as positives. Thus, we do not use the outer product trick for this method. The final objective is $\gL(f) = \gL_1(f) + \gL_2(f)$.

\clearpage

\section{Experimental Details}
\label{appendix:hparams}

We implemented contrastive RL and the baselines using the ACME RL library~\citep{hoffman2020acme} in combination with JAX~\citep{jax2018github}. Precisely, we took the SAC agent\footnote{\url{https://github.com/deepmind/acme/tree/master/acme/agents/jax/sac}} and made the modifications below. Unless otherwise mentioned, we used the same hyperparameters as this implementation.
\begin{enumerate}
    \item Implemented environments that returned observations that contained the original observation concatenated with the goal. While the goals are resampled when sampling from the replay buffer, assuming that observations include the goals means that the Q-function and policy networks do not need to include an additional input for the goal.
    \item Modified the replay buffer to use trajectories rather than transitions. This allows us to sample $(s_t, a_t, s_f)$ triplets.
    \item Modified the critic network to be parametrized as an inner product between state-action representations and goal representations.
    \item Modified the critic loss based on Alg.~\ref{alg:contrastive}. Note that the actor loss does not need to be modified (except for removing the entropy term for state-based tasks).
\end{enumerate}
We summarize the hyperparameters in Table~\ref{tab:hparams}.
Both the state-action encoder and the goal encoder are fully-connected neural networks with 2 layers of size 256 with ReLU activations. We found that normalizing the final representation, applying a final activation function, or using a learnable temperature hurt performance, so we do not use these tricks. For image-based tasks, observations have size $(64, 64, 3)$, and we use a CNN encoder from prior work~\citep{mnih2013playing} to encode the observations before passing them to the encoders. The policy has a similar architecture: an image-encoder for image-based tasks, followed by 2 fully connected layers with size 256 and ReLU activations.

\begin{table}[H]
    \centering
    \footnotesize 
    \caption{Hyperparameters for our method and the baselines.}
    \label{tab:hparams}
    \begin{tabular}{p{7cm}|p{5.5cm}} \toprule
        hyperparameter & value \\ \midrule 
        batch size & 256 \\
        learning rate & 3e-4 for all components \\
        discount & 0.99 \\
        actor target entropy & 0 for state-based experiments, \\
        & $-\text{dim}(a)$ for image-based experiments \\
        target EMA term (for TD3 and SAC) & 0.005 \\
        image encoder architecture & Taken from~\citet{mnih2013playing} \\
        image decoder architecture (for auto-encoder and model-based baselines) & Taken from~\citet{ha2018world} \\
        hidden layers sizes (for actor and representations) & (256, 256) \\
        initial random data collection & 10,000 transitions \\
        replay buffer size & 1,000,000 transitions \\
        samples per insert$^1$ & 256 \\
        train-collect interval$^2$ & 16 for state-based tasks, 64 for image-based tasks \\
        representation dimension ($\dim(\phi(s, a)), \dim(\psi(s_g))$) & 64 \\
        actor minimum std dev & 1e-6 \\
        number of augmentations (for DrQ only) & 4 \\
        logsumexp regularizer coefficient (for CPC only) & 1e-2 \\
        action repeat & None \\
        goals for actor loss & random states (not future states) \\
        \multicolumn{2}{l}{\scriptsize $^1$ How many times is each transition used for training before being discarded.} \\
        \multicolumn{2}{l}{\scriptsize $^2$ We collect $N$ transitions, add them to the buffer, and then do $N$ gradient steps using the experience sampled randomly from the buffer.} \\ \bottomrule
    \end{tabular}
\end{table}

\begin{table}[H]
    \centering
    \footnotesize 
    \caption{Changes to hyperparameters for offline RL experiments (Fig.~\ref{tab:offline-rl}).}
    \begin{tabular}{p{6.7cm}|p{5.5cm}} \toprule
        hyperparameter & value \\ \midrule 
        batch size & $256 \rightarrow 1024$ \\
        representation dimension & $64 \rightarrow 16$ \\
        hidden layers sizes (for actor and representations) & $(256, 256) \rightarrow (1024, 1024)$, as in~\citep{emmons2021rvs}. \\
        goals for actor loss & future states \\ \bottomrule
    \end{tabular}
\end{table}

\subsection{Environments}

\begin{itemize}
    \item[] \textbf{fetch reach} (image, state) -- This task is taken from~\citet{plappert2018multi}. This task involves moving a robotic arm to a goal location in free space. The benchmark specifies success as reaching within 5cm of the goal.
    \item[] \textbf{fetch push} (image, state) -- This task is taken from~\citet{plappert2018multi}. This task involves using a robotic arm to push an object across a table to a goal location. The benchmark specifies success as reaching within 5cm of the goal.
    \item[] \textbf{sawyer push} (image, state) -- This task is taken from~\citet{yu2020meta}. This task involves using a robotic arm to push an object across table to a goal location. The benchmark specifies success as reaching within 5cm of the goal.
    \item[] \textbf{ant umaze} (state) -- This task is taken from~\citet{fu2020d4rl}. This task involves controlling an ant-like robot towards a goal location, which is sampled randomly in a ``U''-shaped maze. Unlike all other tasks in this paper, the goal was lower dimensional than the observation: the goal was just the XY coordinate of the desired position. Following prior work~\citep{chane2021goal, Srinivas2018UniversalPN}, success is defined as reaching within 0.5m of the goal.
    \item[] \textbf{sawyer bin} (image, state) -- This task is taken from~\citet{yu2020meta}. This task involves using a robotic arm to pick up a block from one bin and place it at a goal location in another bin. The benchmark specifies success as reaching within 5cm of the goal.
    \item[] \textbf{point Spiral11x11} (image) -- This task is an image-based version of the 2D navigation tasks in~\citet{eysenbach2019search}. This task involves directly controlling the XY coordinates of an agent to reach a goal in a spiral-shaped maze (see Fig.~\ref{fig:spiral-viz}). We define success as reaching within 2m of the goal.
\end{itemize}

\clearpage
\section{Additional Experiments}
\label{appendix:experiments}

\subsection{Linear regression with the learned features}

\begin{figure}[b]
    \centering
    \begin{subfigure}[b]{0.49\textwidth}
        \centering
        \includegraphics[width=0.6\linewidth]{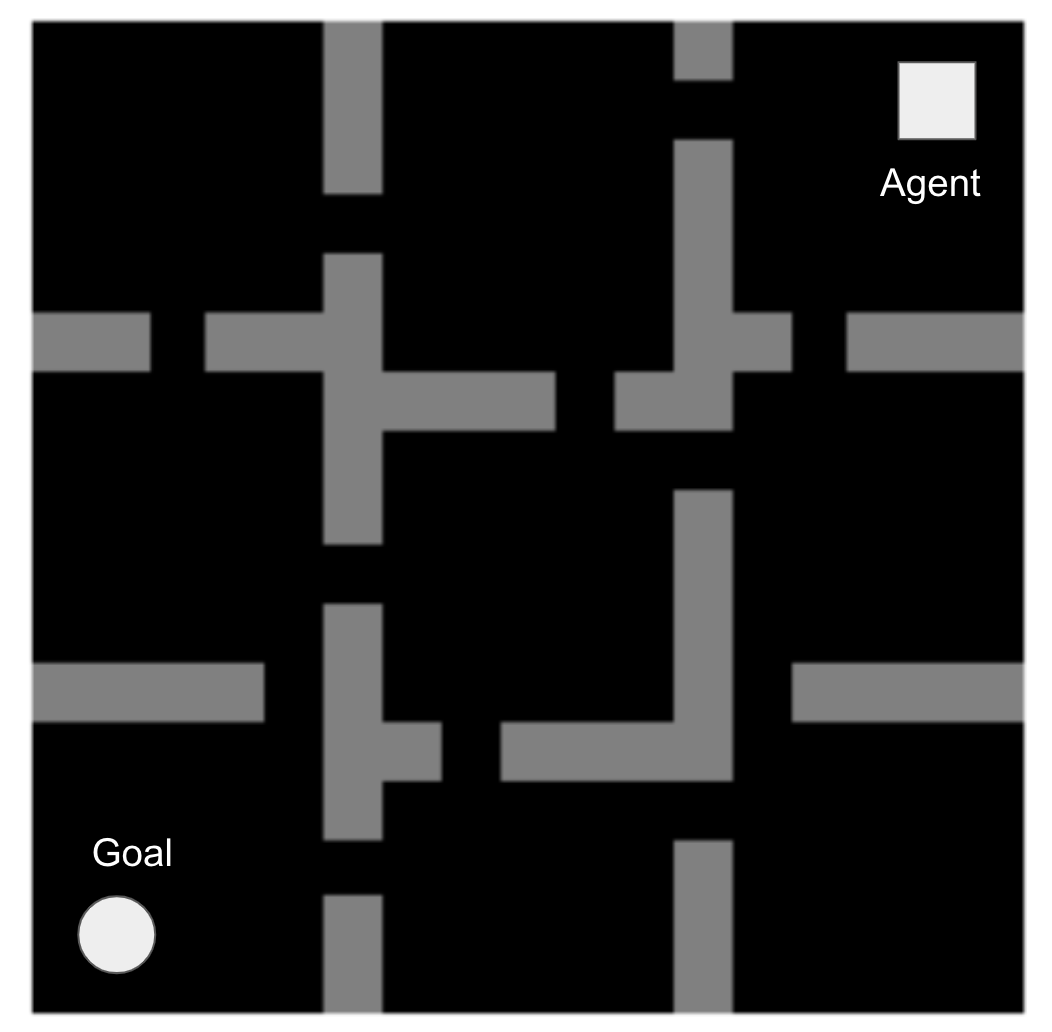}
        \caption{\footnotesize \textbf{Nine-Room environment.}}
        \label{fig:nineroom}
    \end{subfigure}%
    ~
    \begin{subfigure}[b]{0.49\textwidth}
        \includegraphics[width=0.9\linewidth]{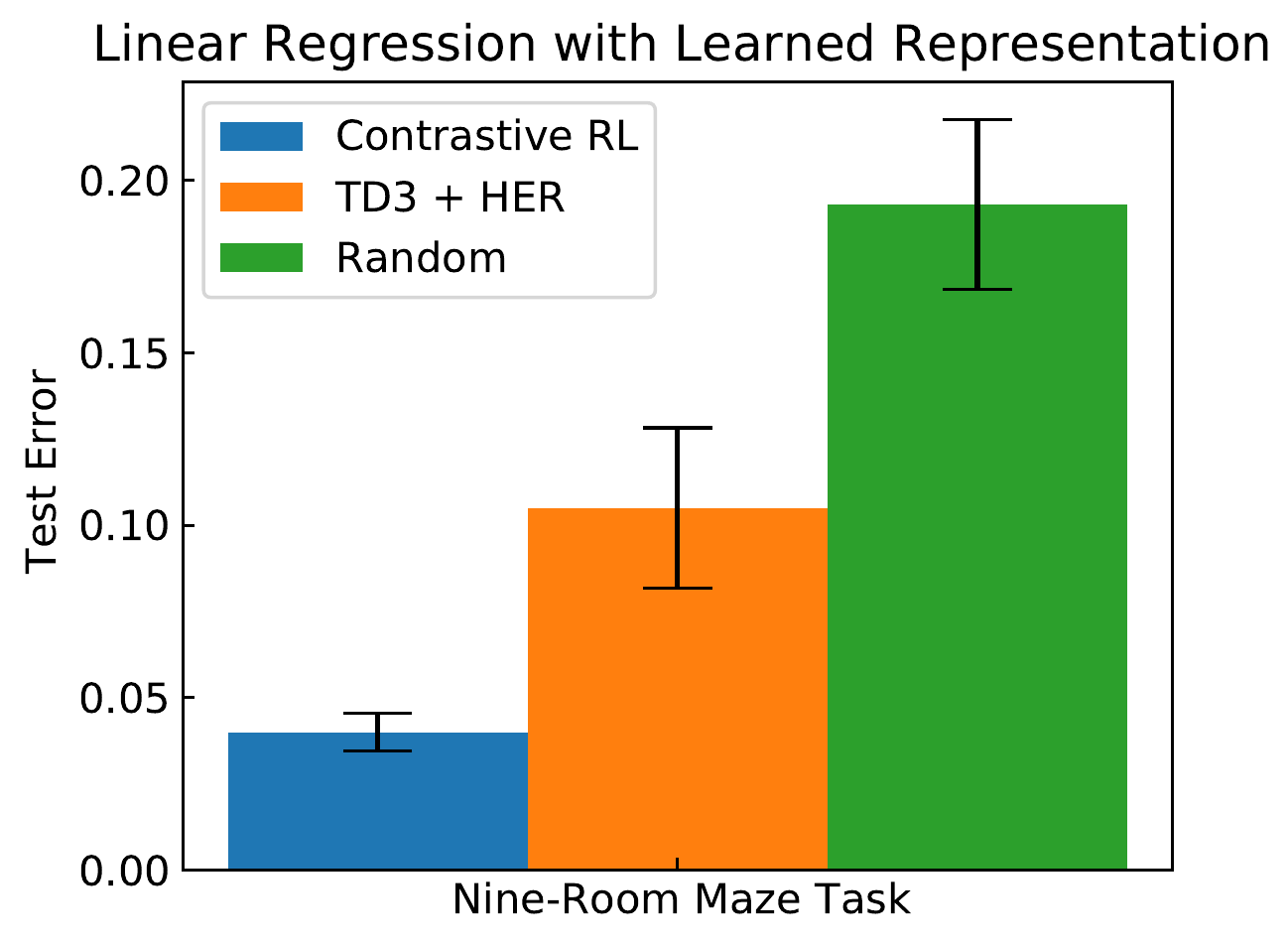}
        \caption{\footnotesize \textbf{Linear probe experiment.}}
    \label{fig:regression}
    \end{subfigure}
    \caption{
    \footnotesize\textbf{Linear regression with the learned features.} Contrastive RL can produce better features for predicting the shortest-path distance, indicating that the learned features have captured highly non-linear information about the environment dynamics.}
\end{figure}

To study the learned representations in isolation we take the state-action representations $\phi(s, a)$ trained on the image-based \texttt{point NineRooms} task, and run a linear probe~\citep{alain2016understanding, han2020self} experiment to see whether the representations have learned to encode task-relevant information (the shortest path distance to the goal).

We use the task of nine-room navigation and run Contrastive RL and TD3+HER on it. We visualize the environment in Fig.~\ref{fig:nineroom} and the agent randomly initialized in one of the nine rooms is commanded to go to the goal position. We dump the replay buffer during training as the dataset and run a linear regression to predict the shortest distance between the agent and the goal. Note that this shortest path distance is not the Euclidean distance since there are walls blocking the way. Fig.~\ref{fig:regression} shows that features learned by contrastive RL can predict this distance better than all baselines.

As shown in Fig.~\ref{fig:regression}, contrastive RL (NCE) learns representations that achieve lower test error than those learned by TD3+HER and by a random CNN encoder.

\subsection{When is contrastive learning better than learning a foreward model?}

\begin{wrapfigure}[14]{R}{0.5\textwidth}
    \centering
    \vspace{-1.5em}
    \includegraphics[width=\linewidth]{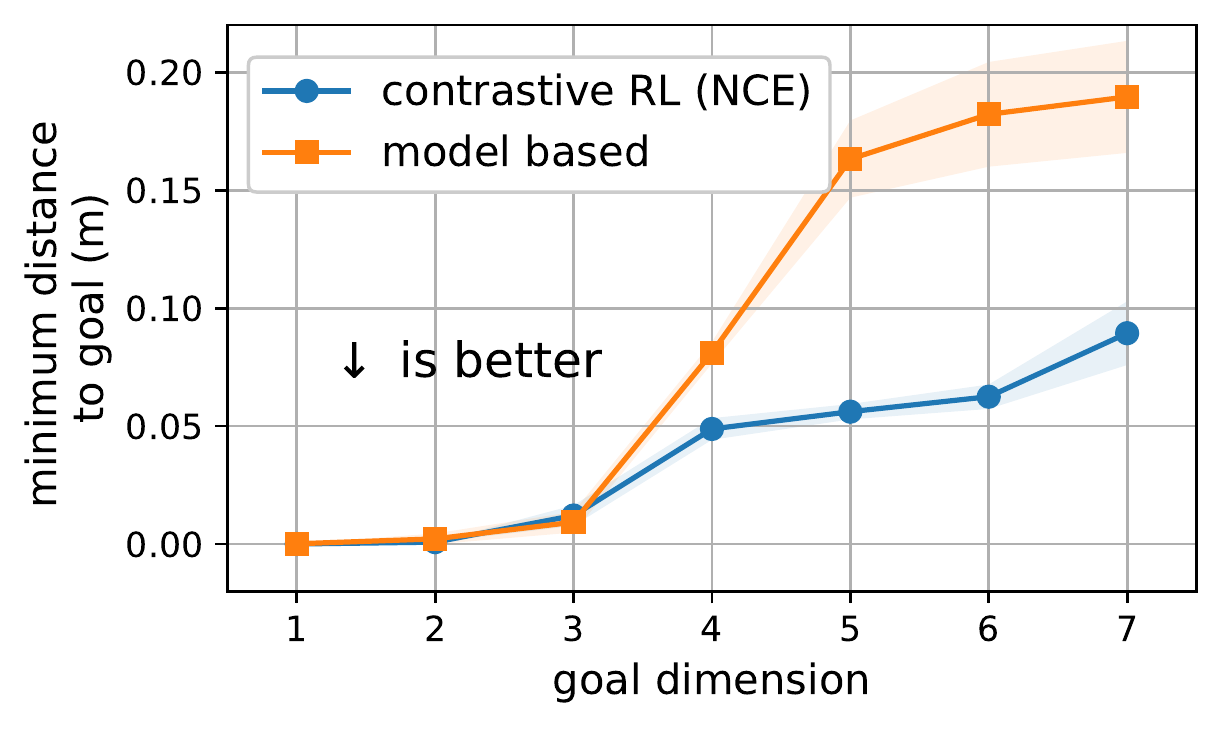}
    \vspace{-1.7em}
    \caption{\footnotesize \footnotesize Contrasive learning outperforms a forward model when the goal is 4-dimensional or larger. Error bars show the standard deviation across 5 random seeds. \label{fig:goal-dim}}
\end{wrapfigure}
In Fig.~\ref{fig:benchmark-state}, we observed that the model-based baseline performed well on the \texttt{ant umaze} task, but poorly on many of the other tasks. One explanation is that the model-based approach will perform well when the goal is relatively low-dimensional, and that contrastive learning will be more useful in settings with higher-dimensional goals. We tested this experiment on the 7-dimensional \texttt{sawyer push} environment. We applied both contrastive RL and the model-based baseline to versions of this task where the goal was varied from 1-dimensional to 7-dimensional. Note that changing the goal dimension changes the task: a 1-dimensional goal corresponds to moving the gripper to the correct X position, whereas a 7-dimensional goal corresponds to moving the object and gripper to the correct poses. We measured the Euclidean distance to the goal ($\downarrow$ is better). We show results in Fig.~\ref{fig:goal-dim}. As expected, higher-dimensional goals are a bit more challenging to achieve. What we are really interested in is the \emph{gap} between the model-based approach and the contrastive RL, which opens up starting with a 4-dimensional goal. Altogether, this experiment provides some evidence that contrastive RL may be preferred over a forward model, even for tasks with very low dimensional goals.

\begin{figure}[ht]
    \centering
    \includegraphics[width=\linewidth]{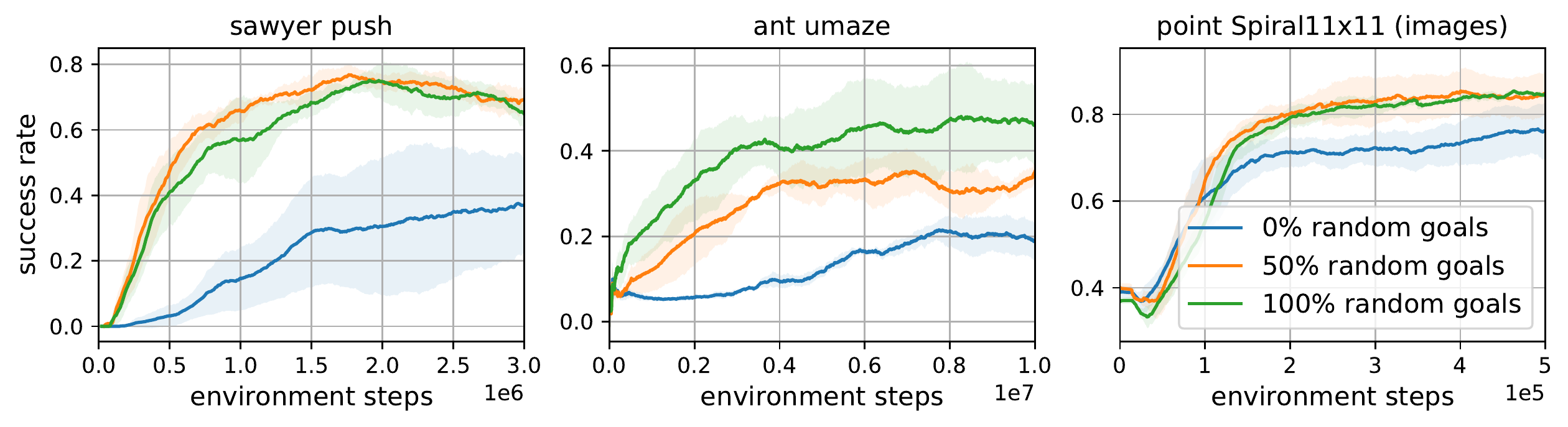}
    \caption{\footnotesize \textbf{Goals used for the actor loss.} Goals are either sampled from the distribution over future states or from a distribution of random states. Error bars show the standard deviation across 5 random seeds.}
    \label{fig:actor-goals}
\end{figure}

\subsection{Goals used in the actor loss}

In theory, the distribution of goals for the actor loss (Eq.~\ref{eq:actor-loss}) does not affect the optimal policy, as long as the distribution has full support. In our experiments, we sampled these goals randomly, in the same way that we sampled negative examples for contrastive learning. We ran an ablation experiment to study this decision, and show results in Fig.~\ref{fig:actor-goals}. These results show that sampling future goals consistently performs poorly, perhaps because it results in only training the policy on how to reach ``easy'' goals. A mixture of future goals and random goals works much better, but the best results seem to come from training on only random goals.

\begin{figure}[ht]
    \centering
    \includegraphics[width=\linewidth]{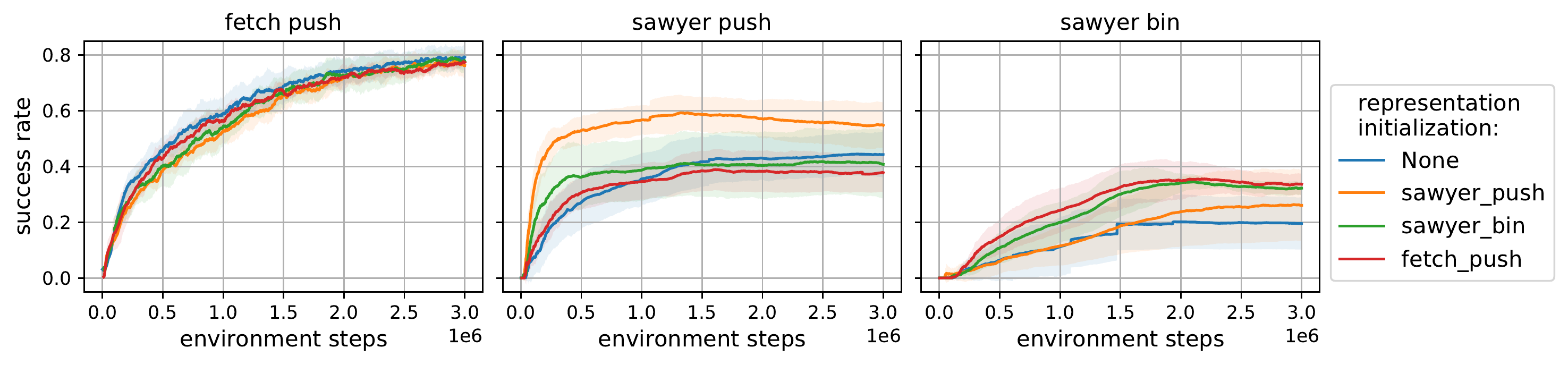}
    \caption{\footnotesize \textbf{Transferring representations to solve new tasks.} After training the representations on one task for ~1M environment steps, we used them to initialize a new agent for solving a new task.}
    \label{fig:contrastive-transfer}
\end{figure}

\subsection{Transferring representations to solve new tasks}

In this experiment, we studied whether the representations learned by contrastive RL (NCE) for one task might be useful for solving another task. We started by training contrastive RL (NCE) on three image-based tasks: \texttt{fetch push}, \texttt{sawyer push}, and \texttt{sawyer bin}. The observations for all tasks look different and the sawyer and fetch tasks have different robots. The two sawyer tasks look the most similar because they both come from the metaworld~\citep{yu2020meta} benchmark suite. We used the representations learned for each of these tasks to initialize a second contrastive RL agent, which we used to solve this same set of tasks. We were primarily interested in transfer -- do the representations from one task help in learning to solve another task? Intuitively, even if the tasks are different, a good representation will capture some structural properties (e.g., identifying the robot arm, and identifying objects), which should transfer across the task.

We show results in Fig.~\ref{fig:contrastive-transfer}. After training on the first task for ~1M environment steps, we used the learned representation as initialization for solving the new task. On the \texttt{fetch push} task, we see little benefit from using pretrained representations, perhaps because the task is relatively easy. On the \texttt{sawyer push}, we see the largest benefit from pretraining the representations on the same task as the target task. More interestingly, we see a small benefit from taking the representations learned on the \texttt{sawyer bin} task and using those to solve the \texttt{sawyer push}. On the most challenging task, \texttt{sawyer bin}, using representations pretrained on either \texttt{fetch push} or \texttt{sawyer bin} can accelerate the solving of this task. Taken together, these results suggest that transferring the representations from one task to another is sometimes useful.

\subsection{Robustness to Environment Perturbations}

\begin{wrapfigure}[6]{R}{0.5\textwidth}
    \centering
    \vspace{-1em}
    \includegraphics[width=0.24\linewidth]{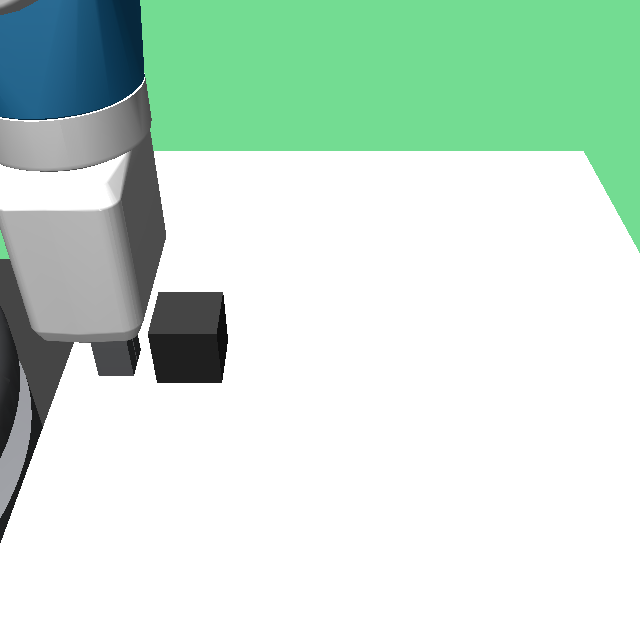}
    \includegraphics[width=0.24\linewidth]{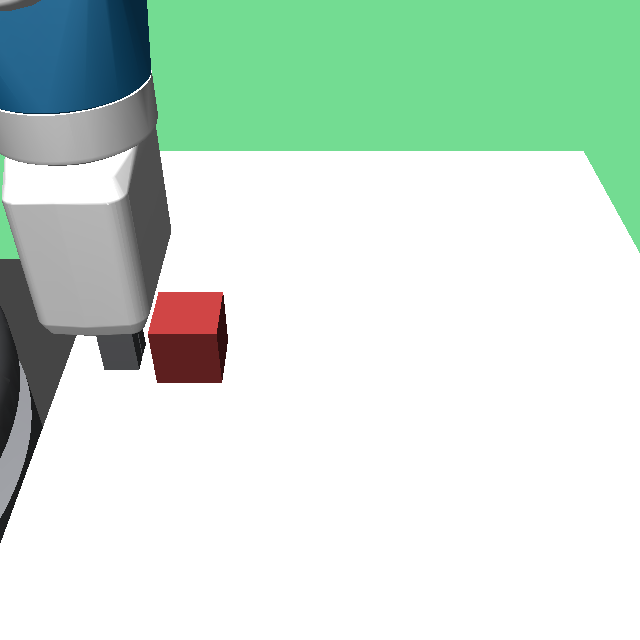}
    \includegraphics[width=0.24\linewidth]{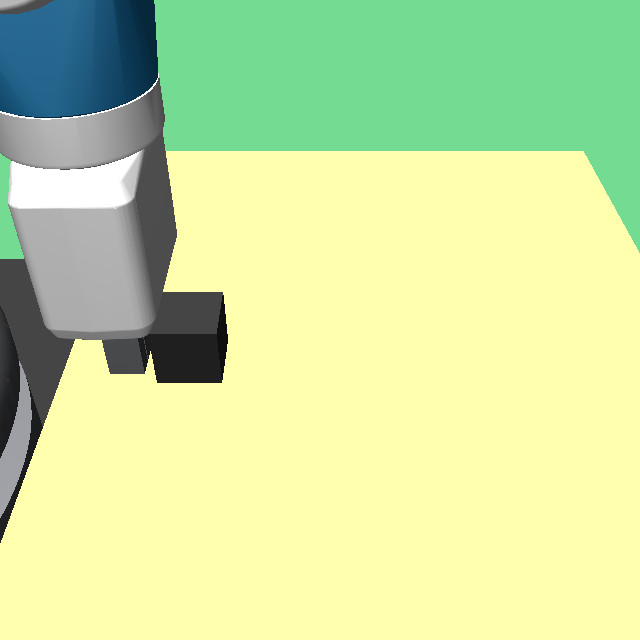}
    \includegraphics[width=0.24\linewidth]{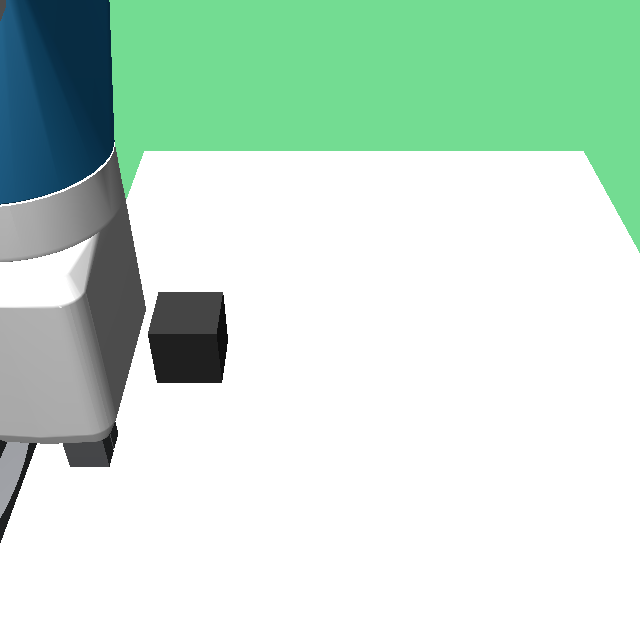}
    \vspace{-1.5em}
    \caption{\footnotesize Perturbations to the image-based \texttt{fetch push} environment.} \label{fig:perturbations}
\end{wrapfigure}
We ran an preliminary experiment to study whether the image-based policies learned by contrastive RL (NCE) were robust to perturbations in the environment. We took an agent trained on the \texttt{fetch push} with image-observations, and evaluated the agent on four variants of the environment (see Fig.~\ref{fig:perturbations}):
\begin{itemize}
    \item Original environment, without modification;
    \item Object color changed from black to red;
    \item Table color changed from white to yellow;
    \item Initial arm position moved towards the camera.
\end{itemize}
In each setting, we evaluate the success rate over 20 trials, and repeated 5 times to compute standard deviations (for a total of 100 trials). The learned agent was robust to the object color, with the success rate changing from $78 \pm5\%$ to $73 \pm10\%$. The agent was also robust to the change in initial position ($87 \pm6\%$). However, changing the table color caused the agent to fail ($0 \pm0\%$), perhaps because the table color consumes a large fraction of the image pixels.

\subsection{Additional figures}

\begin{figure}[ht]
    \centering
    \includegraphics[width=\linewidth]{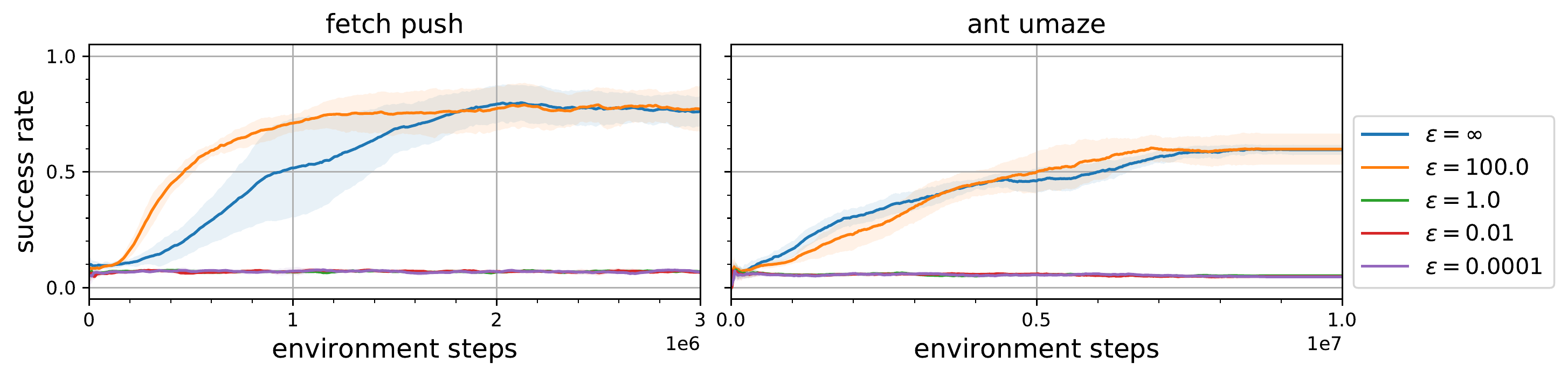}
    \caption{\footnotesize \textbf{Filtered relabeling.} We filter the relabeled experience so that the agent only trains on experience where the probability under the commanded goal is similar to the probability under the actually-reached goal. While such filtering is required to prove convergence, these results suggest that good performance can be achieved without this filtering step.}
    \label{fig:filtering}
\end{figure}

\begin{figure}[th]
    \centering
    \includegraphics[width=\linewidth]{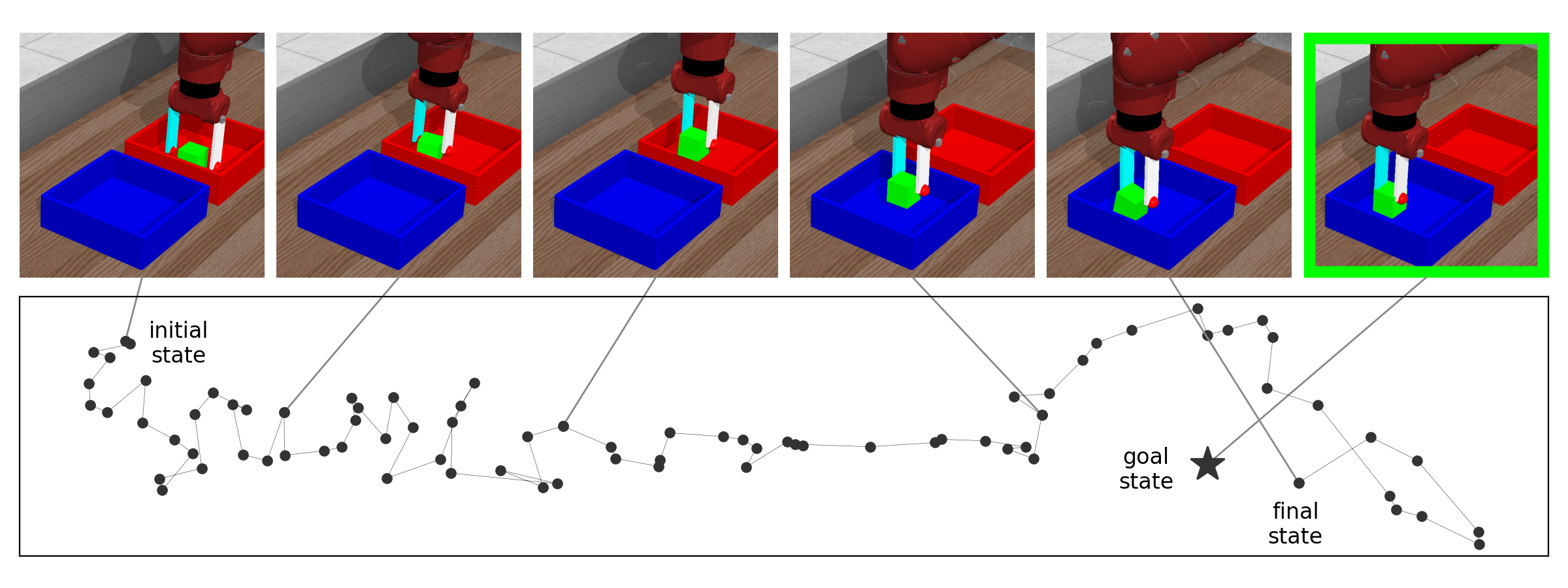}
    \caption{\footnotesize
    \textbf{Visualizing the learned representations.}
    \figtop \; We show five observations from the bin picking task, as well as the goal image.
    \figbottom \; A TSNE embedding of the image representations $\phi(s, a)$ learned by Contrastive RL (NCE). Note that different parts of the task (e.g., reaching, picking, placing) are well separated in the learned representation space.
    }
    \label{fig:bin_repr}
\end{figure}

\begin{figure}
    \centering
    \includegraphics[width=\linewidth]{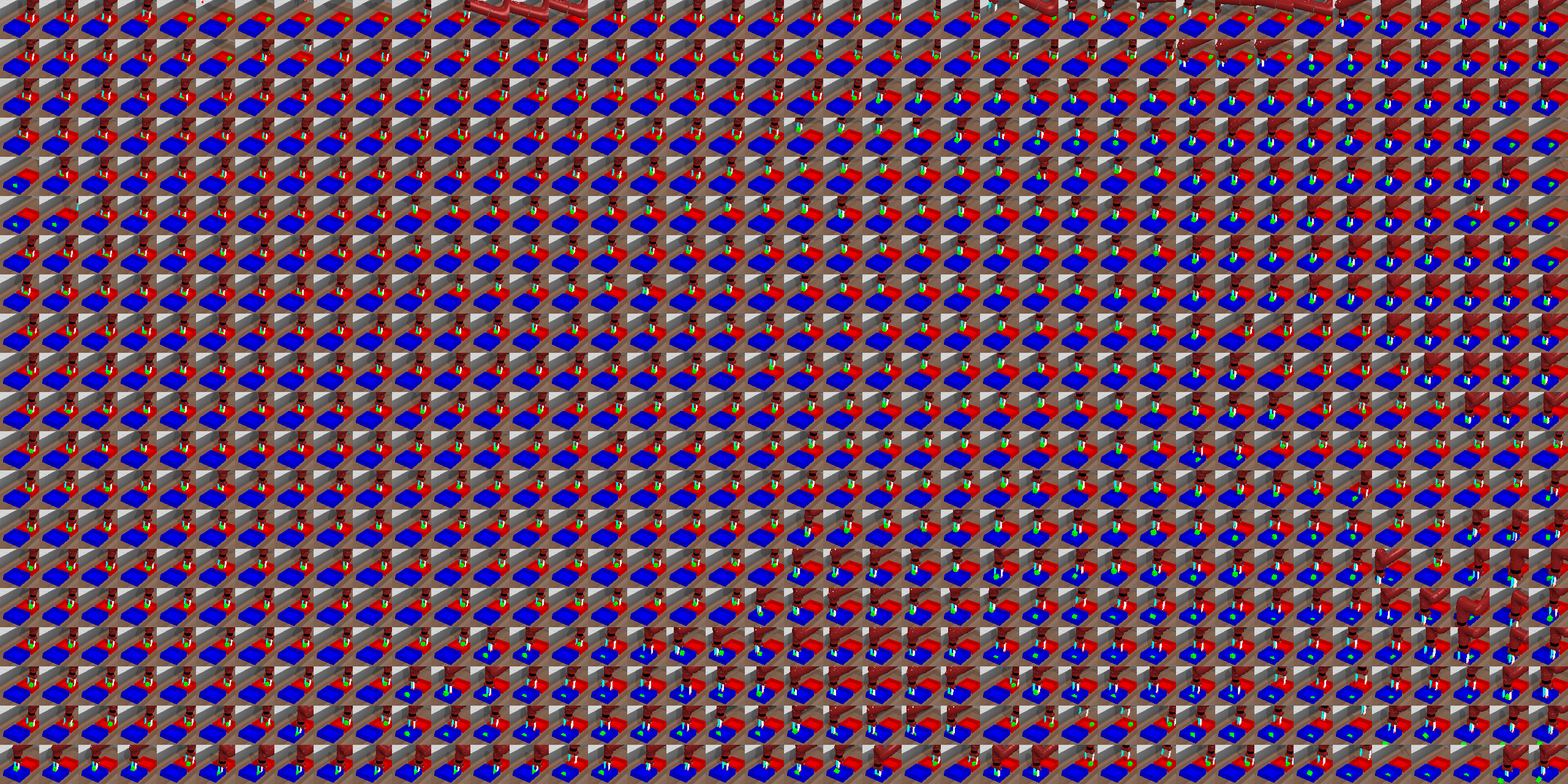}
    \caption{Visualizing the image representations learned by our method on the \texttt{sawyer bin}.
    We compute a TSNE embedding of the representations, and then align the embeddings to a grid using RasterFairy~\citep{klingemann2016raster}.
    }
    \label{fig:raster}
\end{figure}

\begin{figure}
    \centering
    \begin{subfigure}[b]{0.2\textwidth}
    \centering
    \includegraphics[height=3cm]{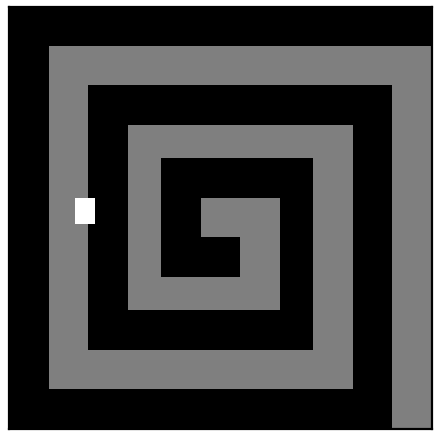}
    \caption{}
    \end{subfigure}
    \begin{subfigure}[b]{0.1\textwidth}
    \hfill
    \end{subfigure}
    \begin{subfigure}[b]{0.2\textwidth}
    \centering
    \includegraphics[height=3cm]{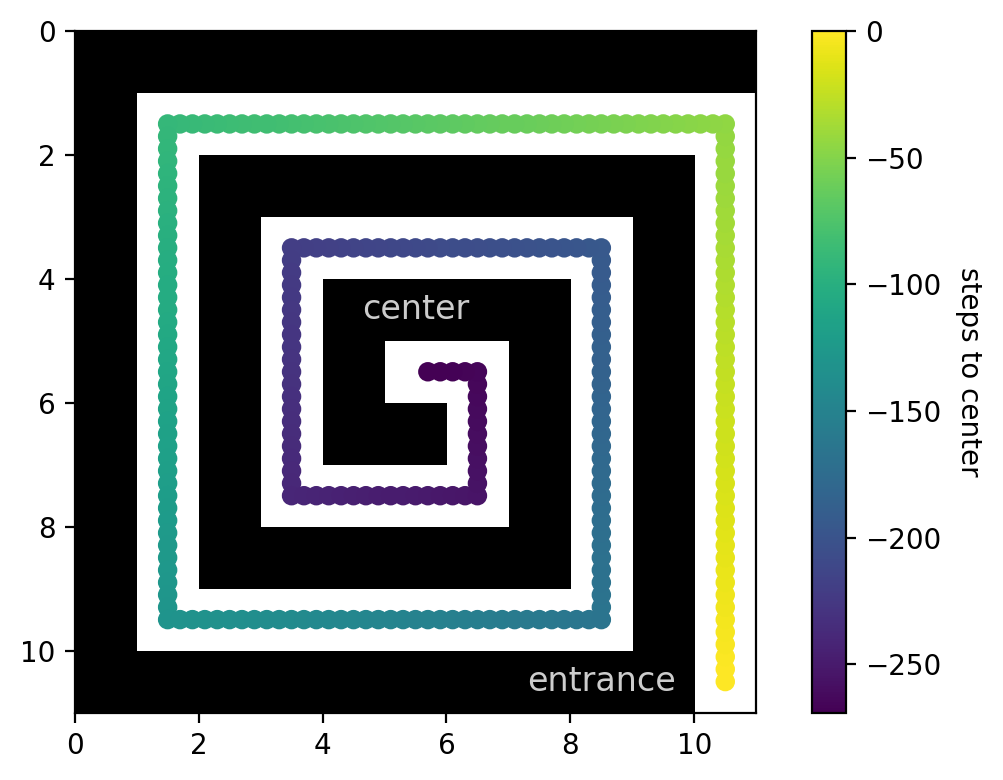}
    \caption{}
    \end{subfigure} \\
    
    \begin{subfigure}[b]{0.3\textwidth}
        \includegraphics[width=\linewidth]{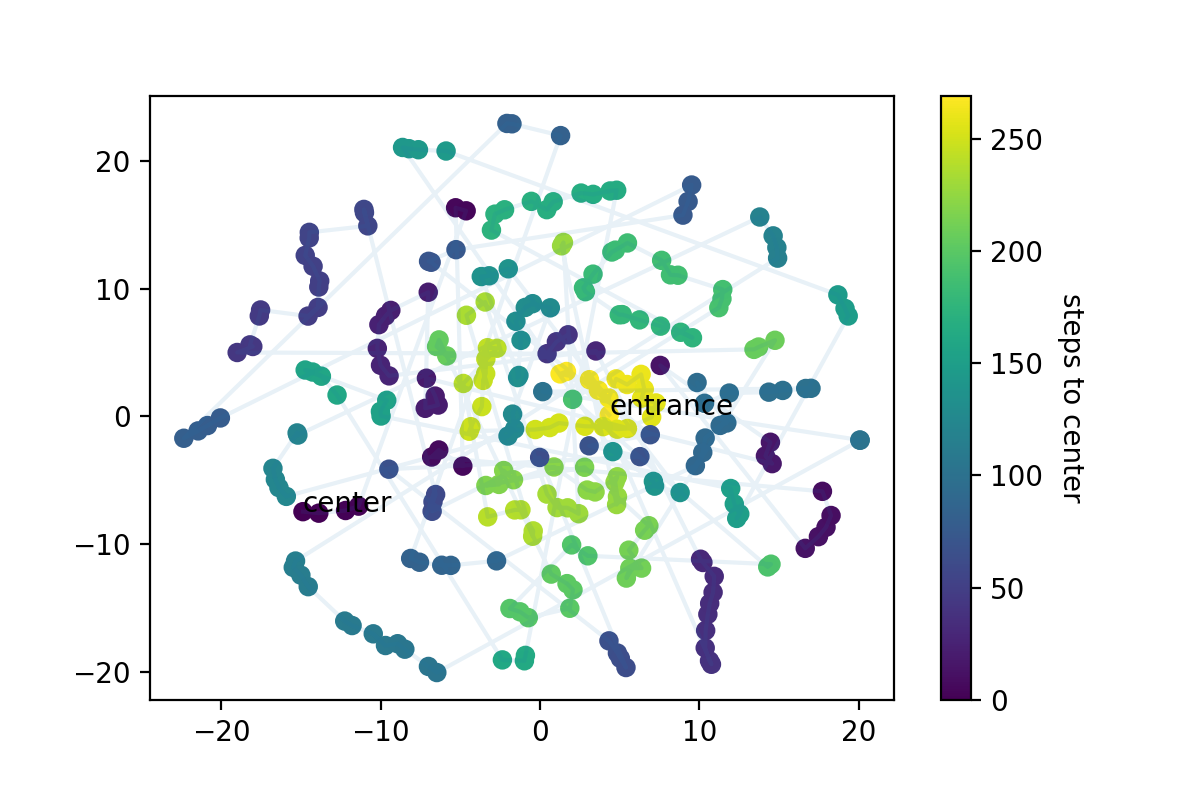}
        \caption{untrained encoder}
    \end{subfigure}%
    ~
    \begin{subfigure}[b]{0.3\textwidth}
        \includegraphics[width=\linewidth]{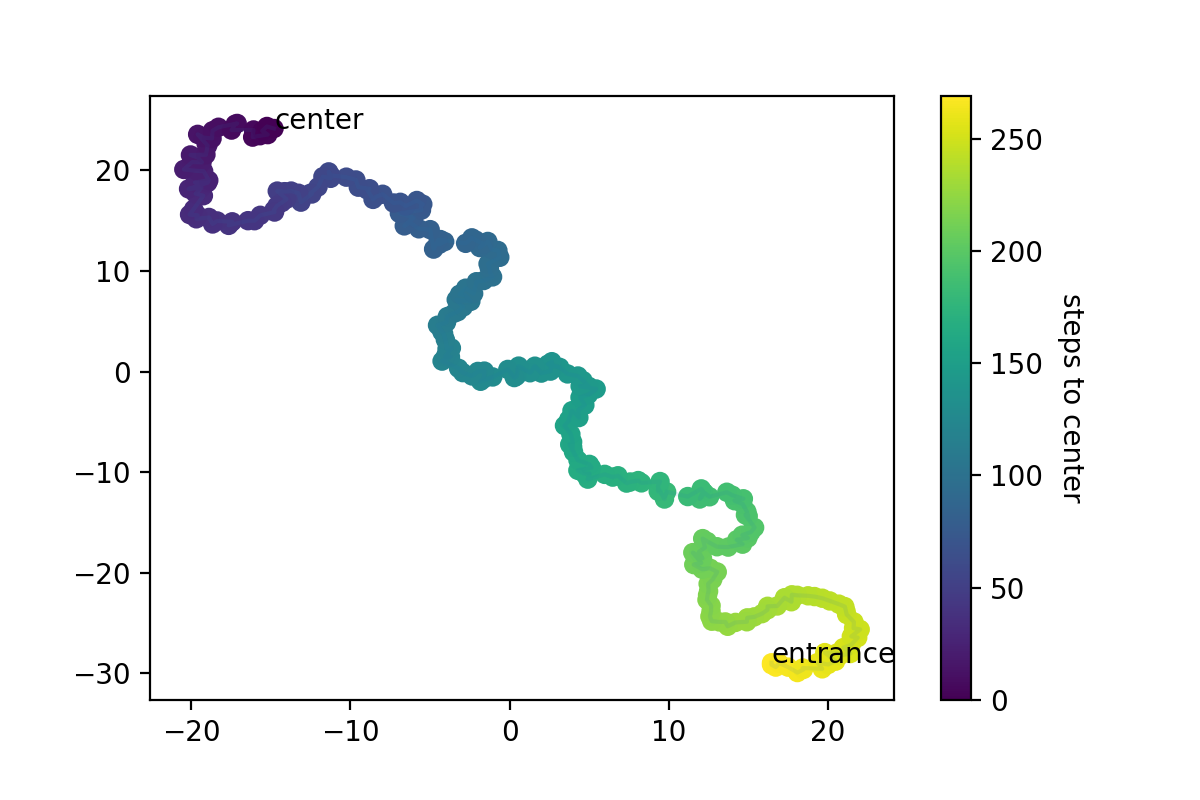}
        \caption{contrastive RL (NCE)}
    \end{subfigure}%
    ~
    \begin{subfigure}[b]{0.3\textwidth}
        \includegraphics[width=\linewidth]{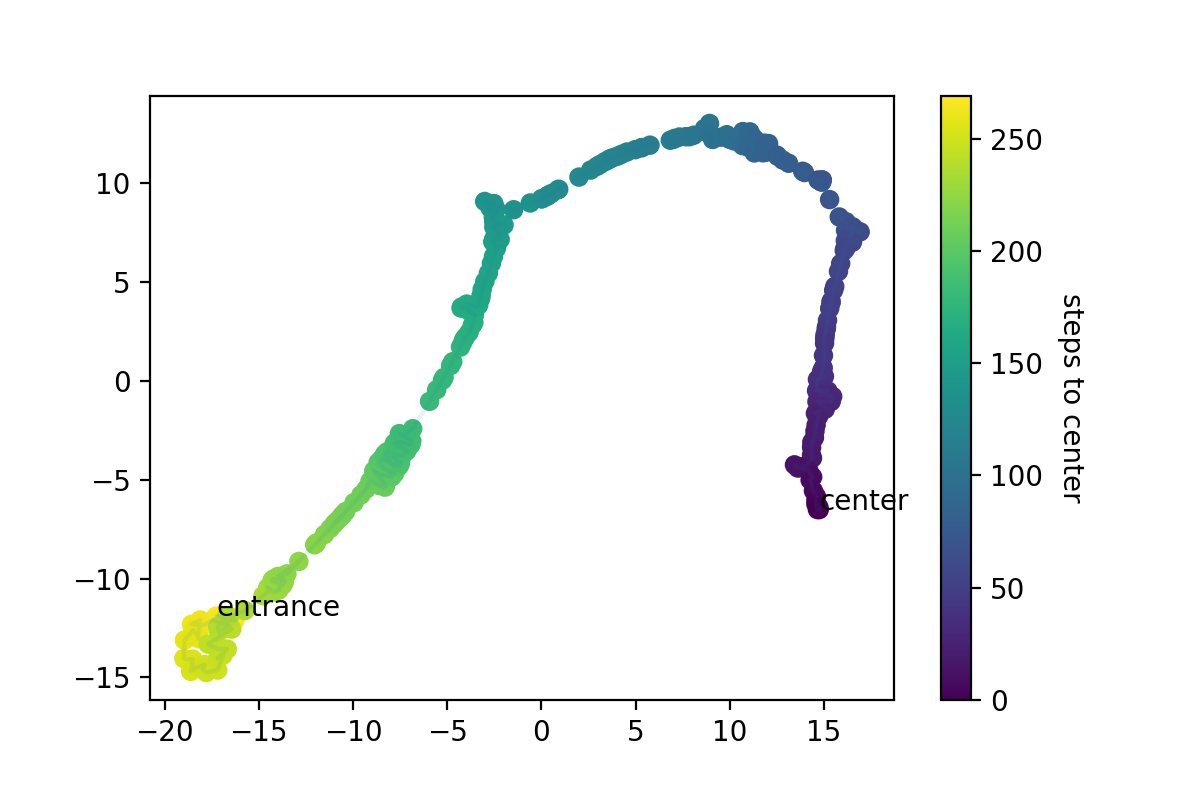}
        \caption{TD3 + HER}
    \end{subfigure}
    \caption{\textbf{TSNE embedding of representations $\phi(s, a)$.}
    \emph{(a)} Using the \texttt{point Spiral11x11} task, \emph{(b)} we generated image observations at 270 locations throughout the maze. We computed the state-action representations of these images, using action = (0, 0). \emph{(c, d, e)} A TSNE embedding of these representations reveals that the untrained encoder does not capture the structure of the environment, whereas both our method and the TD3 + HER baseline do capture the maze structure.}
    \label{fig:spiral-viz}
\end{figure}

\begin{figure}
    \centering
    \begin{subfigure}[c]{0.3\textwidth}
        \centering
        \includegraphics[width=\linewidth]{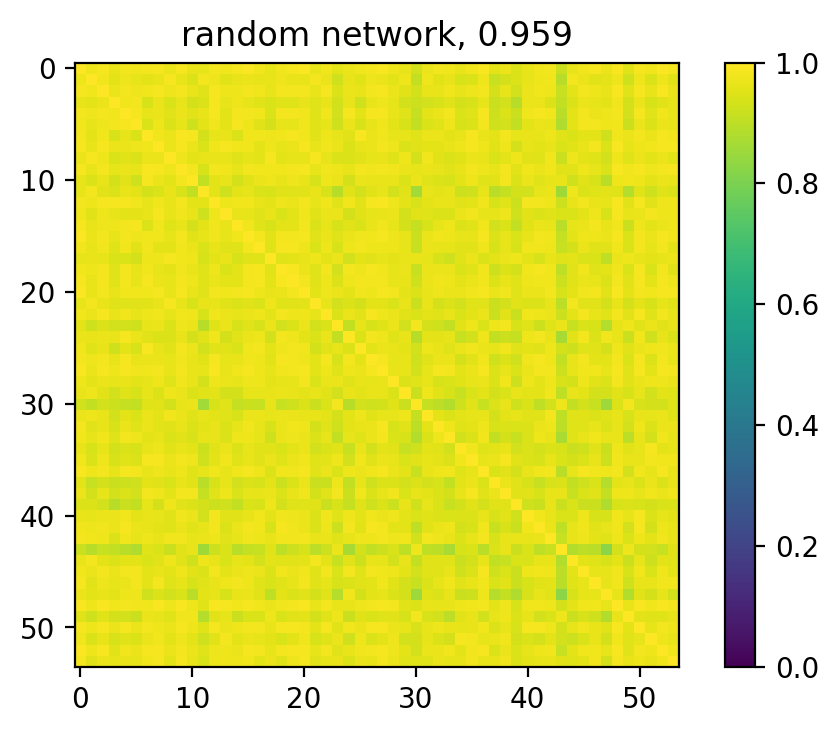}%
        \caption{Random neural network}
    \end{subfigure}
    \begin{subfigure}[c]{0.3\textwidth}
        \centering
        \includegraphics[width=\linewidth]{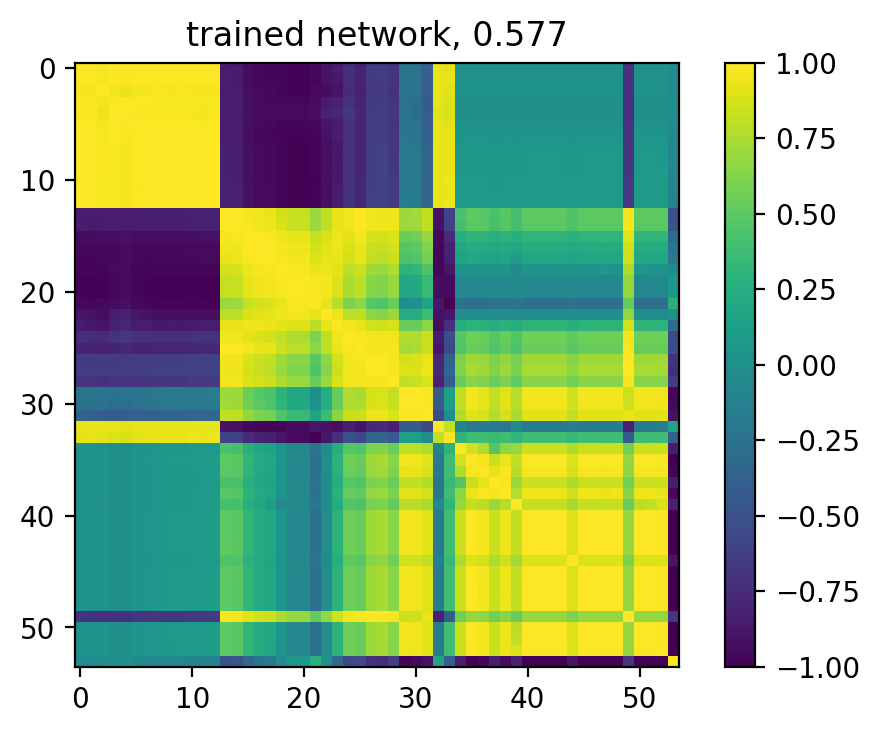}
        \caption{C-learning}
    \end{subfigure}
    \begin{subfigure}[c]{0.3\textwidth}
        \centering
        \includegraphics[width=\linewidth]{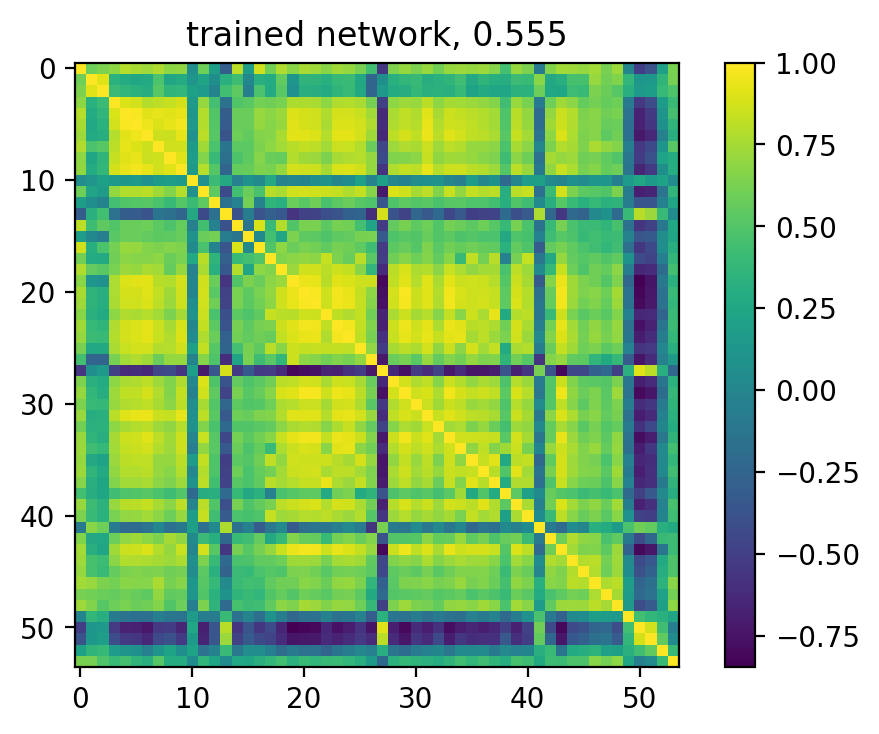}
        \caption{contrastive RL (NCE)}
    \end{subfigure}
    \caption{ \footnotesize
    \textbf{Analyzing the gradients.}
    We plot the cosine similarity between the (normalized) gradients of the critic function with respect to the goal images. An untrained network has high gradient similarity, meaning that updates to one state/task affect the networks predictions at many other states/tasks, a phenomenon that prior work has identified as being detrimental to learning~\citep{kumar2020implicit, yu2020gradient, yang2021overcoming, achiam2019towards}. Our method converges to a network where gradients at one state have a low similarity with gradients at other states. A similar plot showing gradients with various state inputs shows a similar effect.
    }
    \label{fig:aliasing}
\end{figure}

This section presents additional figures.
\begin{itemize}
    \item Fig.~\ref{fig:filtering} compares contrastive RL (NCE) with varying values of the filtering parameter $\epsilon$, described in Sec.~\ref{sec:analysis}.
    \item Fig.~\ref{fig:bin_repr} -- This plot shows a TSNE embedding of the state-action representations $\phi(s, a)$ for one trajectory of the \texttt{bin picking} task. This experiment uses image observations.
    \item Fig.~\ref{fig:raster} -- This plot shows a TSNE embedding of the state-action representations from the same \texttt{bin picking} task. We sampled states and actions using a trained agent. After computing the TSNE embedding, we used RasterFairy~\citep{klingemann2016raster} to rectify the embeddings to a grid.
    \item Fig.~\ref{fig:spiral-viz} -- A TSNE embedding of image representations from the \texttt{point Spiral11x11} task.
    \item Fig.~\ref{fig:aliasing} -- Using the same representations for the \texttt{point Spiral11x11} task, we measure the similarity between the critic gradients when evaluated at the same state but different goals, $\langle \frac{\partial f}{\partial s}\vert_{(s, g)}, \frac{\partial f}{\partial s}\vert_{(s, g')}\rangle$.
\end{itemize}

\clearpage
\section{Failed Experiments}
\label{appendix:failed}

\begin{enumerate}
    \item \emph{Representation normalization}: We experimented with many ways of normalizing the learned representations, including L2 normalization, scaling the representations by a learned temperature parameter, and applying an elementwise tanh activation. None of these modifications consistently improved performance.
    \item \emph{Momentum encoder}: Prior contrastive learning methods have found it useful to use a target encoder or momentum buffer. We experimented with many similar tricks, including using a momentum buffer for goal representations, sampling some fraction of goal representations from a fixed random distribution, increasing the learning rate for the state encoder's final layer, and decreasing the learning rate for all layers in the goal encoder. None of these tricks consistently improved performance.
    \item \emph{Frame stacking} -- This tended to decrease performance slightly.
    \item \emph{Loss scaling} -- Our contrastive RL (NCE) method uses the negative label much more often than the positive label. We tried scaling the loss terms so that the negative and positive examples received the same total weight, but found this had no effect on performance.
\end{enumerate}

\clearpage

\end{document}